\PassOptionsToPackage{table}{xcolor} %
\documentclass[10pt,twocolumn,letterpaper]{article}

\usepackage[pagenumbers]{cvpr} %

\usepackage{cuted} %

\usepackage{currfile} %

\usepackage{caption} %

\makeatletter
\@ifpackageloaded{lineno}{%
  \@ifpackagelater{lineno}{2026/02/11}{%
    \@ifpackagelater{lineno}{2026/02/17}{}{%
      \RemoveFromHook{build/column/before}[lineno]%
      \AddToHook{build/column/before}{%
        \protected@write\@auxout{}{%
          \string\def\string\@LN@column{\if@firstcolumn2\else1\fi}%
        }%
      }%
    }%
  }{}%
}{}
\makeatother

\usepackage{amsmath,amsfonts,bm}
\usepackage{xcolor}

\def\eqref#1{equation~\ref{#1}}

\def\1{\bm{1}}

\DeclareMathAlphabet{\mathsfit}{\encodingdefault}{\sfdefault}{m}{sl}
\SetMathAlphabet{\mathsfit}{bold}{\encodingdefault}{\sfdefault}{bx}{n}

\usepackage{xcolor}
\usepackage{colortbl}
\definecolor{oursrow}{RGB}{232,241,252}
\newsavebox{\cdtabbox}
\usepackage{graphicx}
\usepackage{tikz}
\usetikzlibrary{positioning,arrows.meta,calc,fit,backgrounds,decorations.pathreplacing}

\usepackage{booktabs}
\usepackage{tabularx}
\usepackage{array}
\usepackage{multirow}
\usepackage{makecell}
\usepackage{adjustbox}
\usepackage{pifont}
\newcolumntype{C}{>{\centering\arraybackslash}X}

\setcitestyle{square,comma,numbers,sort&compress}

\usepackage{amsmath,amsfonts,amssymb}
\usepackage{bm}

\usepackage{enumitem}
\setlist[itemize]{leftmargin=*}

\usepackage{caption}
\usepackage{subcaption}
\AtBeginDocument{%
  \crefname{figure}{Fig.}{Figs.}%
  \crefname{table}{Tab.}{Tabs.}%
  \crefname{section}{Sec.}{Secs.}%
}

\usepackage{xspace}

\definecolor{cvprblue}{rgb}{0.21,0.49,0.74}
\usepackage{fontawesome5}
\usepackage[most]{tcolorbox}
\usepackage[T1,OT1]{fontenc} %
\definecolor{cdbg}{HTML}{FCFCFA}\definecolor{cdink}{HTML}{161B26}\definecolor{cdink2}{HTML}{3A4152}
\definecolor{cdmuted}{HTML}{6B7185}\definecolor{cdred}{HTML}{FF4D5E}\definecolor{cdaccent}{HTML}{4F8CFF}
\definecolor{cdorange}{HTML}{F2731C}\definecolor{cdpink}{HTML}{EE4062}\definecolor{cdmauve}{HTML}{C0589A}\definecolor{cdviolet}{HTML}{7C6FF0}
\IfFileExists{T1Inter-TLF.fd}
  {}
  {}
\newcommand{\cdhy}{\kern-.06em-\kern-.06em}
\IfFileExists{T1ComicNeue-TLF.fd}{\newcommand{\cdhand}{\fontencoding{T1}\fontfamily{ComicNeue-TLF}\selectfont}}{\newcommand{\cdhand}{\fontencoding{T1}\fontfamily{phv}\selectfont}}
\usetikzlibrary{decorations.pathmorphing,patterns}
\tikzset{sketch/.style={decorate,decoration={random steps,segment length=6pt,amplitude=.5pt},rounded corners=2pt,line width=.7pt}}

\newcommand{\cdpill}[2]{\href{#2}{\tikz[baseline=(n.base)]{\pgfmathsetseed{3}\node[inner xsep=5pt,inner ysep=3pt,fill=white](n){\cdhand\bfseries\footnotesize\color{cdink}#1\,$\nearrow$};
  \draw[sketch,cdink!70] (n.south west) rectangle (n.north east);}}}
\usepackage{ragged2e}
\usepackage[pagebackref,breaklinks,colorlinks,allcolors=cvprblue]{hyperref}
\makeatletter
\def\cvprsection{\@startsection {section}{1}{\z@}
   {-10pt plus -2pt minus -2pt}{7pt} {\large\bf\color{cvprblue}}}
\def\cvprsubsection{\@startsection {subsection}{2}{\z@}
   {-8pt plus -2pt minus -2pt}{5pt} {\elvbf\color{cvprblue}}}
\def\cvprsubsubsection{\@startsection {subsubsection}{3}{\z@}
   {-6pt plus -2pt minus -2pt}{3pt} {\tenbf\color{cvprblue}}}
\makeatother

\def\confName{CVPR}
\def\confYear{2027}

\title{CrashDojo: Accident-Grounded World Models\\as Closed-Loop Safety Examiners for Autonomous Driving Planners}

\author{
Bowen Jing$^{1*}$\quad
Haoyun Tang$^{2*}$\quad
Ruiyang Hao$^{3}$\quad
Chenchen Ge$^{4}$\quad
Weitao Zhou$^{5}$\quad
Haibao Yu$^{6\dagger}$\\[2pt]
{\normalsize $^{1}$Tuojing AI\quad
$^{2}$Xi'an Jiaotong University\quad
$^{3}$King's College London}\\
{\normalsize $^{4}$Southeast University\quad
$^{5}$Beijing Simple AI Technology\quad
$^{6}$The University of Hong Kong}\\[2pt]
{\small $^{*}$Equal contribution\quad $^{\dagger}$Corresponding author}
}

\hypersetup{pdfauthor={Bowen Jing, Haoyun Tang, Ruiyang Hao, Chenchen Ge, Weitao Zhou, Haibao Yu},pdftitle={CrashDojo: Accident-Grounded World Models as Closed-Loop Safety Examiners for Autonomous Driving Planners}}
\begin{document}
\twocolumn[{%
\vspace*{-0.5in}
\begin{tcolorbox}[enhanced,width=\textwidth,frame hidden,boxrule=0pt,arc=0pt,colback=cdbg,
  left=7mm,right=7mm,top=5.5mm,bottom=5mm,
  overlay={\pgfmathsetseed{8126}
    \draw[sketch,cdink!75] ([xshift=1pt,yshift=-1pt]frame.north west) rectangle ([xshift=-1pt,yshift=1pt]frame.south east);
    \draw[sketch,cdink!30] ([xshift=2.6pt,yshift=-2.2pt]frame.north west) rectangle ([xshift=-.4pt,yshift=2.4pt]frame.south east);
}]
\raggedright\color{cdink}
{\cdhand\fontsize{8.5}{10}\selectfont\color{cdmuted}{\color{cdred}$\bullet$}\enspace Autonomous driving\enspace$\cdot$\enspace Safety evaluation\par}
\vspace{1.4mm}
{\cdhand\bfseries\fontsize{38}{41}\selectfont Crash{\color{cdorange}D}{\color{cdpink}o}{\color{cdmauve}j}{\color{cdviolet}o}\par}
\vspace{2mm}
{\cdhand\fontsize{14}{17}\selectfont\color{cdink2}Accident-Grounded World Models as Closed-Loop\\Safety Examiners for Autonomous Driving Planners\par}
\vspace{4mm}
{\cdhand\bfseries\small Bowen Jing$^{1*}$\quad Haoyun Tang$^{2*}$\quad Ruiyang Hao$^{3}$\quad Chenchen Ge$^{4}$\quad Weitao Zhou$^{5}$\quad Haibao Yu$^{6\dagger}$\par}
\vspace{1.2mm}
{\cdhand\fontsize{8.5}{11}\selectfont\color{cdmuted}
$^{1}$Tuojing AI\quad $^{2}$Xi'an Jiaotong University\quad $^{3}$King's College London\quad $^{4}$Southeast University\\
$^{5}$Beijing Simple AI Technology\quad $^{6}$The University of Hong Kong\qquad $^{*}$Equal contribution\quad $^{\dagger}$Corresponding author\par}
\vspace{3mm}
\tikz{\pgfmathsetseed{27}\draw[sketch,cdink!40] (0,0) -- (\linewidth,0);}\par
\vspace{2.5mm}
{\small\justifying\noindent{\cdhand\bfseries Abstract.} Evaluating autonomous-driving planners requires safety-critical traffic interactions that are rarely encountered in recorded data. Guided traffic generators can induce collisions and adapt to ego responses, but a collision objective alone does not fully specify the interaction being tested: which actor initiates the conflict, what maneuver it performs, and how strongly it intervenes. Accident evidence can inform these choices, yet closed-loop execution introduces a further challenge: fixed trajectories cannot adapt to the ego response, while unconstrained replanning may alter the intended maneuver and its timing. We present \textbf{CrashDojo}, which connects accident-grounded interaction selection to persistent request-guided control of a frozen traffic world model. Road structure, participant roles, and accident evidence rank candidate actor--mechanism requests, fixing the selected actor, maneuver parameters, and timing before rollout. Geometry-specific objectives guide target motion on the original request clock, while contextual motion references and observed ego--target proximity modulate contact pressure during execution. Candidate checks assess executable trajectory prefixes. Across five planners on 100 nuScenes and 100 nuPlan scenes, CrashDojo matches or exceeds SAFE-SIM in initial-target contact in three of four dataset--planner groups while reducing target--background contact and target body off-road rates in all four. Motion-distribution comparisons and a separate three-seed component study further characterize the generated encounters.\par}
\vspace{4mm}
\begin{minipage}[c]{0.42\linewidth}
\cdpill{Project page}{https://crashdojo.github.io}\hspace{2.5mm}\cdpill{Code}{https://github.com/HaoyunT/CrashDojo}
\end{minipage}\hfill
\begin{minipage}[c]{0.57\linewidth}\raggedleft
\raisebox{-.5\height}{\includegraphics[width=6.5717cm]{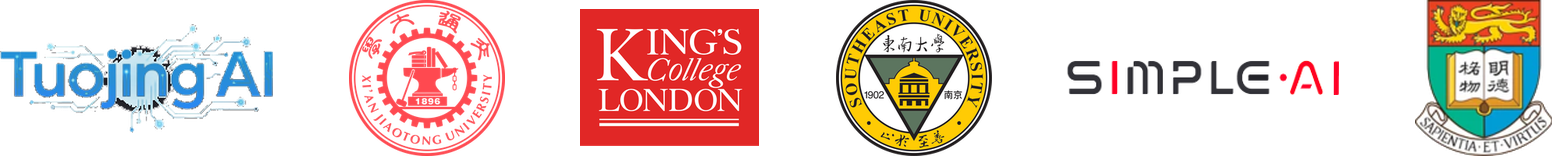}}
\end{minipage}
\end{tcolorbox}
\vspace{5mm}
}]

\section{Introduction}
\label{sec:intro}

\begin{figure*}[t]
\centering
\includegraphics[width=.92\textwidth]{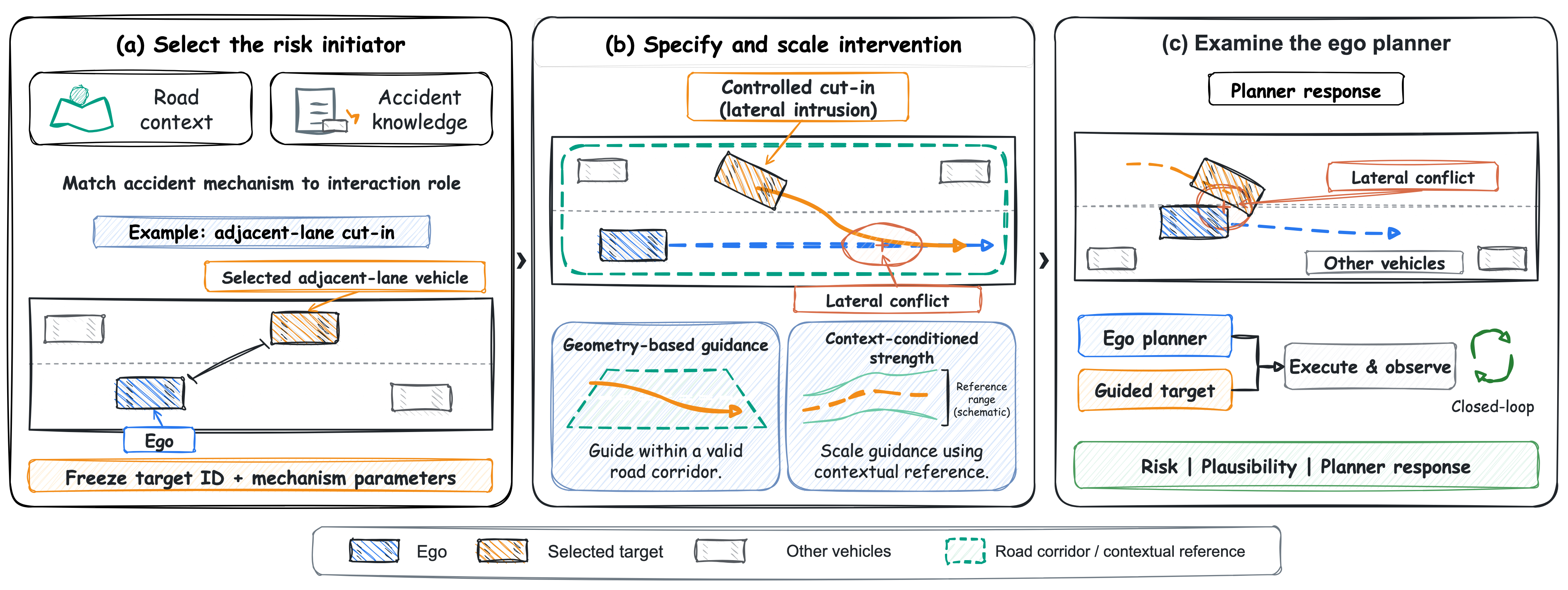}
\caption{\textbf{From accident context to a planner challenge.} (a) Select an actor and fix its mechanism/request parameters. (b) Guide its motion using pair geometry and contextual strength. (c) Execute the interaction and observe the ego response. Relative positions update throughout execution. Scenes, trajectories, and reference ranges are schematic.}
\label{fig:concept}
\end{figure*}

Improving autonomous-driving planners requires tests that expose failures under rare but recognizable traffic interactions. Recorded driving data provide limited coverage of such encounters, while generative models can synthesize such scenarios~\cite{jiang2024scenediffuser,wu2024smart}. Guided generation further directs these models toward safety-critical situations~\cite{rempe2022generating,chang2024safesim,lin2025causal}. In closed-loop evaluation, a traffic world model can serve as an \emph{interactive safety examiner}: it constructs challenging traffic and updates as ego planner responds. \Cref{fig:concept} shows this process with an adjacent-lane cut-in.

Existing methods have demonstrated collision-oriented guidance and interactive traffic generation. SAFE-SIM combines collision-type proposals with closed-loop adaptation~\cite{chang2024safesim}, while CCDiff incorporates causal interaction structure into guided generation~\cite{lin2025causal}. These approaches establish the ability to generate challenging encounters, but a collision objective alone does not fully specify the interaction being tested. A lead vehicle braking and an adjacent-lane vehicle cutting in may both produce contact, yet they demand different responses from the ego. Constructing a targeted challenge therefore involves not only inducing an adverse outcome, but also determining which actor should initiate the conflict and what maneuver it should perform.

Closed-loop execution further complicates this specification. As the ego brakes or yields, the relative geometry changes, and a fixed target trajectory may no longer produce the intended encounter. Conversely, redefining the maneuver at each replanning step can alter its timing and progression. A useful distinction is therefore between the interaction being requested and the trajectory generated to realize it: the former should remain specified while the latter adapts to the ego response. This also raises the question of how to choose the interaction in the first place. Pre-crash typologies relate dangerous interactions to road conditions and participant roles~\cite{najm2007precrash}, while driving logs provide contextual references for maneuver intensity.

\looseness=-1 We present \textbf{CrashDojo}, which connects accident-informed interaction selection to adaptive closed-loop traffic generation. Road structure, participant roles, and accident evidence rank candidate actor--mechanism requests, fixing the selected actor, maneuver parameters, and reference clock before rollout. Geometry-specific objectives guide a frozen Scene Transformer and flow-matching traffic model~\cite{ngiam2022scene,lipman2023flow}, while contextual motion references and observed ego--target proximity modulate contact pressure. Candidate checks assess the trajectory prefix to be executed. The request remains fixed as generated motion adapts to the ego response. We evaluate four generators across five planners on nuScenes and nuPlan, measuring initial-target contact, target--background contact, road adherence, and motion distributions. Relative to SAFE-SIM, CrashDojo matches or exceeds target contact in three of four dataset--planner groups while reducing target--background contact and target body off-road rates in all four (Tabs.~\ref{tab:main_nuPlan}--\ref{tab:main_nuScenes}; \cref{fig:two_axis}). A separate three-seed study examines the selection policy and mechanism-specific control bundle. Our contributions are as follows:
\begin{figure}[t]
\centering
\includegraphics[width=\linewidth]{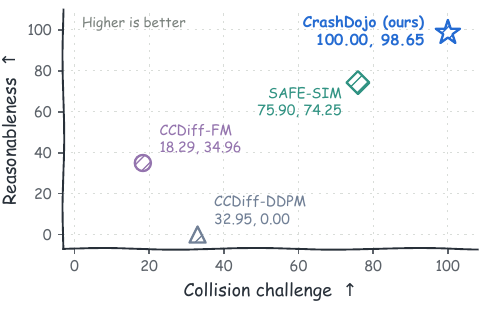}
\caption{\textbf{Collision challenge vs.\ reasonableness} (higher is better; 200 scenes, five planners). Each axis averages per-metric min--max scores over the four methods: challenge uses Ego CR, scene CR, and TTC at $-0.5$\,s; reasonableness uses nine collateral-contact and off-road metrics. Scores are relative to the compared methods; raw values are in \cref{sec:extended}.}
\label{fig:two_axis}
\vspace{-4mm}
\end{figure}

\begin{itemize}[nosep]
\clubpenalty=10000
\widowpenalty=10000
\brokenpenalty=10000
\item \looseness=-1 \textbf{Accident-grounded challenge construction.} We use road structure, participant roles, and accident evidence to select explicit actor--mechanism requests with bounded maneuver parameters before rollout.

\item \textbf{Persistent request-guided control.} We preserve maneuver parameters and timing across replanning while adapting target motion through geometry-specific objectives and context-dependent contact pressure.

\item \looseness=-1 \textbf{Targeted challenge evaluation.} Across two datasets and five planners, we evaluate target contact, collateral effects, and motion distributions, supported by a separate three-seed component study.
\end{itemize}

\begin{table*}[t]
\centering
\caption{\textbf{Design choices in safety-critical traffic generation.} Knowledge lists explicitly specified structural sources; None indicates no separate source. Penalties and checks describe behavioral controls, without implying guaranteed validity.}
\label{tab:mechanisms}
\begingroup
\small
\setlength{\tabcolsep}{4pt}
\renewcommand{\arraystretch}{1.05}
\begin{tabularx}{\textwidth}{>{\raggedright\arraybackslash}p{.11\textwidth}
 >{\raggedright\arraybackslash}p{.16\textwidth}
 >{\raggedright\arraybackslash}p{.19\textwidth}
 >{\raggedright\arraybackslash}p{.23\textwidth}
 >{\raggedright\arraybackslash}X}
\toprule
Method & Knowledge & Intervention & Behavioral controls & Interaction\\
\midrule
SAFE-SIM~\cite{chang2024safesim} & None & Trajectory; collision type & Motion prior; regularization & Closed-loop adaptation\\
CCDiff~\cite{lin2025causal} & Causal graph & Joint trajectories & Realism constraints & Policy rollout\\
KG-ASG~\cite{wang2026kgasg} & Collision knowledge & Primary/support actors & Rule and physics checks & Feedback retries\\
STRIVE~\cite{rempe2022generating} & None & Latent trajectories & Motion prior; bicycle model & Ego response\\
\midrule
\rowcolor{oursrow}\textbf{CrashDojo} & Pre-crash typologies & Actor; mechanism & Motion and road penalties & Persistent request; replanning\\
\bottomrule
\end{tabularx}
\endgroup

\end{table*}

\section{Related work}
\label{sec:related}
\textbf{Predictive and interactive world models.}
Driving world models represent futures through video~\cite{gao2024vista}, occupancy~\cite{zheng2024occworld}, or agent motion. Agent-based models make vehicle interactions directly accessible through joint behavior generation~\cite{suo2021trafficsim}, scene and city-scale rollout~\cite{jiang2024scenediffuser,tan2025scenediffuser}, and motion-token prediction~\cite{wu2024smart}. Scene Transformer provides a multi-agent motion representation~\cite{ngiam2022scene}, while CTG and ProSim introduce temporal-logic and prompt-based control~\cite{zhong2023guided,tan2024promptable}. These advances supply traffic futures on which targeted interventions can operate. Planner evaluation additionally requires connecting generated traffic to an execution environment. DriveArena combines traffic management with image synthesis~\cite{yang2025drivearena}; Bench2Drive-R separates behavior from rendering~\cite{you2024bench2driver}, and NeuroNCAP uses neural rendering for safety tests~\cite{ljungbergh2024neuroncap}. WorldLens and interPlan examines world-model fidelity and long-tail planner generalization~\cite{liang2026worldlens,hallgarten2024vehicle}. \textit{CrashDojo} operates at the traffic-state interface, specifying the actor and maneuver that a frozen generator is asked to enact during planner interaction. 

\looseness=1 \noindent \textbf{Safety-critical scenario generation.} STRIVE searches a learned traffic prior; KING and AdvSim optimize kinematic or sensor-consistent perturbations~\cite{rempe2022generating,hanselmann2022king,wang2021advsim}. SAFE-SIM combines test-time diffusion guidance, collision-type trajectory proposals, and closed-loop adaptation~\cite{chang2024safesim}. CCDiff incorporates causal interaction structure into diffusion guidance~\cite{lin2025causal}; CounterScene uses interaction-critical actor selection for spatial and temporal counterfactual guidance~\cite{jing2026counterscene}. These works already establish guided, interactive traffic generation. Knowledge can also shape the scenario before trajectory search. ChatScene retrieves description--code knowledge for language-driven construction~\cite{zhang2024chatscene}, while Any2Critical and CrashAgent incorporate crash-related context~\cite{huang2026any2critical,li2025crashagent}. \textit{CrashDojo} carries such context into an executable request: pre-crash typologies and road-conditioned statistics~\cite{najm2007precrash,nhtsa_crss} order concrete actor--mechanism candidates, and the selected parameters determine the objectives used throughout replanning. The distinction is the connection between evidence-based construction and persistent control within the generator. \Cref{tab:mechanisms} compares representative designs.

\noindent \textbf{Feasibility and interpretation of tests.} Feasibility and responsibility impose distinct requirements. STRIVE searches for a collision-free ego solution after constructing a collision~\cite{rempe2022generating}; SaFeR restricts token resampling using a learned feasible region~\cite{cui2026safer}; CARS uses careful-and-competent reference drivers~\cite{xiao2026learning}; StyleDrive studies driving preferences using scene-conditioned behavior annotation~\cite{hao2026styledrive}. KG-ASG assigns primary and supporting actors with rule and physics checks~\cite{wang2026kgasg}. Formal constraints clarify behavioral assumptions~\cite{censi2019liability,maierhofer2020formalization,shalevshwartz2017formal}, but failure to find a safe response alone does not establish inevitability~\cite{fraichard2004inevitable}. SafeBench organizes standardized safety evaluation~\cite{xu2022safebench}, and action-conditioned predictions need not recover an observed counterfactual~\cite{zhang2026driving}. WM-Cov distinguishes requested, realized, and valid evidence~\cite{cui2026wmcov}. \textit{CrashDojo} likewise retains the requested mechanism alongside measurements of the executed interaction. Our main comparison evaluates complete systems under common scenes and planners, while the component study matches inputs within each group to examine selection and guidance.

\section{Problem formulation}
\label{sec:problem}
At simulation time $t$, state $s_t$ contains the map, observed histories, and current vehicle states. An ego planner $\pi$ controls vehicle $e$; the generator controls a selected target $a$. We represent a challenge request as
\begin{equation}
 r=(s_0,a,m,\eta,\mathcal W,\Omega),
 \label{eq:request}
\end{equation}
where $m$ is the accident mechanism, $\eta$ its geometry, timing, and intensity parameters, $\mathcal W$ the outer intervention window that bounds every timing entry of $\eta$, and $\Omega$ its motion and road limits. The target identity, mechanism, $\eta$, and $\mathcal W$ remain fixed for the whole rollout. Pair geometry $g_t^a$, including relative pose and path relation, updates with the scene.

A frozen traffic model proposes target trajectories conditioned on $s_t$. Guidance uses current observations and the ego proposal $\widehat\tau_t^e=\pi(s_t)$; the simulator executes a short prefix and repeats. Offline request preparation scores nominal conflict against the logged ego continuation (\cref{sec:knowledge_details}), which is not supplied as future ego states during online guidance. Context $c$, fixed at initialization alongside the request, supplies a driving-data reference for intervention strength.

The request defines the intended interaction, whereas its outcome depends on the executed trajectories and the ego response. We therefore distinguish contact with the selected target, collateral contact, road adherence, and motion distributions in the evaluation. The construction record identifies the challenge that was specified; the rollout measurements characterize the encounter that was produced.

\section{The CrashDojo framework}
\label{sec:method}
\begin{figure*}[t]
\centering
\includegraphics[width=\textwidth]{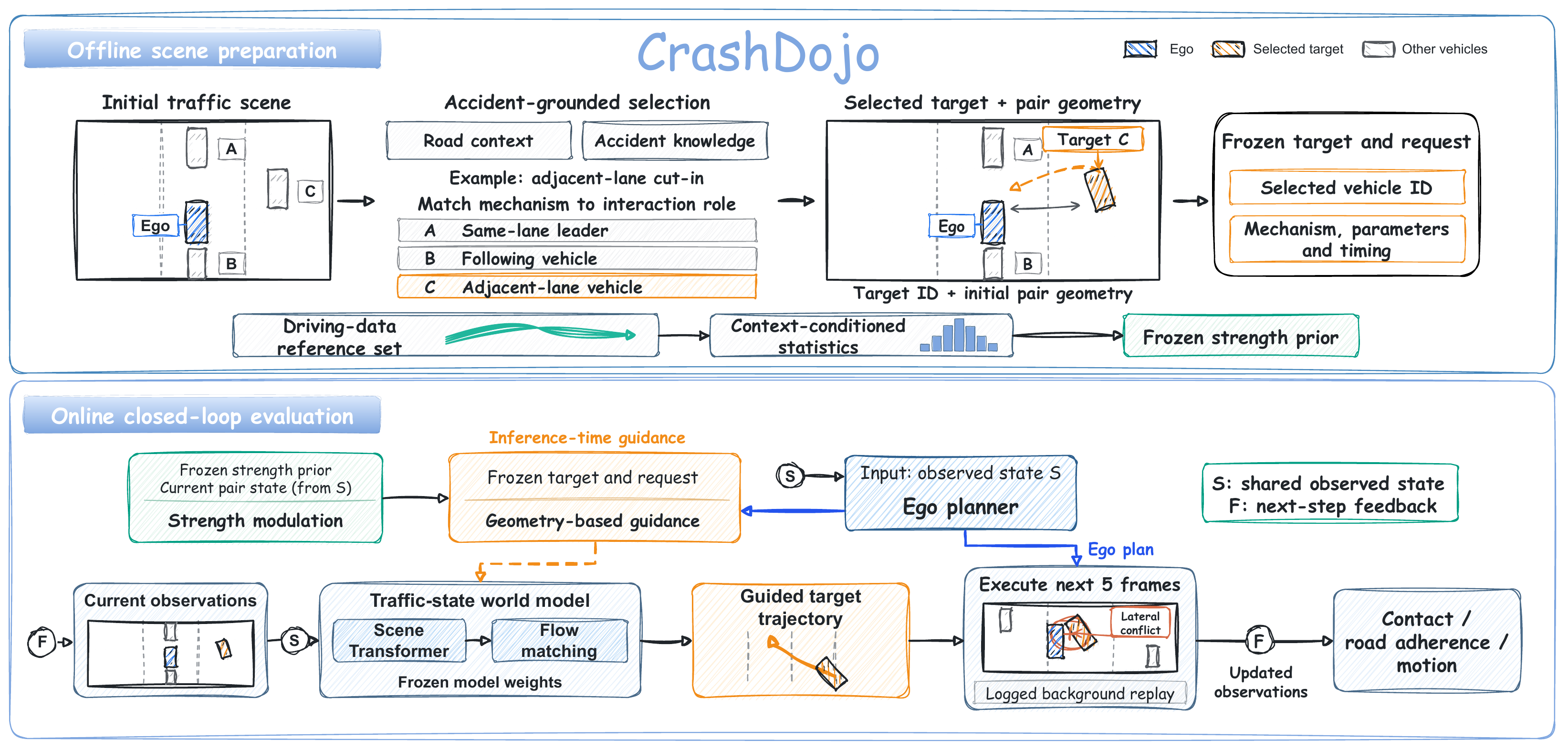}
\caption{\textbf{The CrashDojo framework.} Offline preparation fixes the actor--mechanism request and contextual prior. Online, current observations and the ego proposal inform geometric guidance; observed relative motion modulates contact pressure. Every 0.5\,s, the simulator executes five frames and returns new observations, while the request and contextual prior persist. A--C identify candidate vehicles. All depicted scenes and distributions are schematic.}
\label{fig:pipeline}
\end{figure*}

CrashDojo implements the persistent request of \cref{sec:problem} through the pipeline in \cref{fig:pipeline}. Offline preparation determines the target actor, mechanism, maneuver parameters, and their supporting evidence. These quantities remain fixed during evaluation. At each replanning step, a frozen traffic model proposes target trajectories, mechanism-specific objectives guide the proposals relative to the current ego plan, and contextual feedback adjusts contact pressure. Candidate prefix checks then determine which trajectory segment is executed. The request thus connects challenge construction to its subsequent closed-loop execution.

\subsection{Accident-grounded actor--mechanism selection}
\label{sec:knowledge}\label{sec:selection}
A nearby actor may be unable to initiate a relevant conflict. We enumerate tuples $\xi=(a,m,\eta)$ using a road--accident lookup informed by pre-crash typologies~\cite{najm2007precrash}. Lead braking requires a same-corridor relation; cut-in and intrusion require an adjacent receiving lane; crossing requires intersecting routes. Each tuple receives seven checks covering traffic situation, road structure, accident support, vehicle capability/history, nominal conflict, maneuver intensity, and configuration completeness. Each check records pass, fail, or insufficient evidence; when subchecks are combined, a failure takes precedence over missing evidence. Unavailable support and a contradicted condition therefore remain distinguishable in the request record. Mapping examples and exact checks appear in \cref{sec:knowledge_details}.

The selector orders the enumerated candidate pool $\mathcal C(s_0)$ lexicographically; each $R$ is an ordinal level where 0 is best:
\begin{align}
 \xi^\star&=\operatorname*{arg\,min}_{\xi\in\mathcal C(s_0)}^{\rm lex}\mathbf k(\xi),\nonumber\\
 \mathbf k(\xi)&=(N_{\rm fail},N_{\rm unk},R_{\rm acc},R_{\rm conflict},\nonumber\\
 &\hspace{10mm}R_{\rm intensity},\bar d_{\min},\bar A_{\rm lon},\mathrm{ID}(a)).
 \label{eq:core_selection}
\end{align}
\looseness=-1 Failed and unknown checks take priority, followed by the ordinal evidence, proximity, and intensity tiers $R_{\rm acc}$, $R_{\rm conflict}$, $R_{\rm intensity}$; the rounded minimum body gap $\bar d_{\min}$ and longitudinal effort $\bar A_{\rm lon}$ resolve finer differences, and the actor ID is a semantics-free tie-break (\cref{sec:impl_selection}). A geometric score therefore cannot compensate for weaker evidence in an earlier criterion. Nominal conflict checks are offline; selection retains the chosen actor, mechanism, parameters, and full verdict record. The nuScenes main cohort mixes native-map and fallback request geometry (\cref{sec:extended}); nuPlan and the development cohort use native-map requests only.

\subsection{Geometry-specific intervention}
\label{sec:guidance}
At absolute prediction time $T_k$, let $d_k$ be the target's lateral offset on its stored route and $v_k$ its speed. Mechanism references specify $d_k^\star$ and $v_k^\star$ on the persistent request clock. Longitudinal requests change speed while retaining lateral offset; cut-ins specify a lateral transition; intrusion requests add boundary-relative depth and return phases. Crossing and merging additionally specify route progress and encounter timing. Replanning updates the predicted motion without restarting these references.

For example, a cut-in with lateral increment $\Delta d$ uses a smooth transition $S(u)=10\bar u^3-15\bar u^4+6\bar u^5$, with $\bar u=\operatorname{clip}(u,0,1)$:
\begin{align}
 d_k^\star&=d_0+\Delta d\,S((T_k-t_0)/T_{\rm lc}),\nonumber\\
 v_k^\star&=v_0+(v_{\rm goal}-v_0)S((T_k-t_v)/T_v).
 \label{eq:core_reference}
\end{align}
The onsets $t_0,t_v$, durations $T_{\rm lc},T_v$, increment $\Delta d$, and speed goal $v_{\rm goal}$ are entries of $\eta$ bounded by $\mathcal W$; $d_0,v_0$ are the target's offset and speed at the request anchor. Relative positions still change as the ego responds. This separation preserves the requested maneuver across replanning while allowing its trajectory to adapt to the ego response.

The action term combines normalized lateral/speed tracking with applicable phase-completion, boundary, and arrival penalties. The next five frames receive additional tracking weight because they are executed before the next replan. To create an encounter, let $D_k$ be the signed separating-axis gap between the target and ego boxes: $D_k>0$ means separation, and $D_k\leq0$ means overlap. Over active mechanism indices $\mathcal I_t$, contact pressure minimizes
\begin{equation}
 L_{\rm contact}=\min_{k\in\mathcal I_t}
 \left\{[D_k+\epsilon]_+^2/4+\rho(T_k-t)\right\},
 \label{eq:core_contact}
\end{equation}
where $[x]_+=\max(x,0)$, $T_k$ is the absolute prediction time, $\epsilon=0.05$\,m, and $\rho=1$ in seconds-equivalent scaling: a margin-adjusted 2\,m gap costs as much as 1\,s of delay. The delay term favors earlier contact; cut-in pressure activates halfway through the requested transition and remains within $\mathcal W$. Crossing requests add a conflict-region occupancy cost $L_{\rm cross}$ (\cref{sec:implementation}); we denote the augmented objective by $L_{\rm pressure}=L_{\rm contact}+0.1L_{\rm cross}$.

The full target cost is
\begin{equation}
 \begin{split}
 L_t={}&w_m L_{\rm action}+\lambda_t L_{\rm pressure}
 +0.1L_{\rm prior}+30L_{\rm limits}\\
 &+L_{\rm deviation}+100L_{\rm background}.
 \end{split}
 \label{eq:core_cost}
\end{equation}
Here $w_m=5$ for longitudinal requests and 20 otherwise, and $\lambda_t$ is the contact-pressure coefficient of \cref{sec:calibration}; $L_{\rm limits}$ penalizes violations of $\Omega$. $L_{\rm prior}$ penalizes changes to the model-proposed acceleration and yaw rate. $L_{\rm deviation}$ limits departure from the first model proposal within a replan, with soft average/final displacement budgets of 2/4\,m measured against that proposal rather than the recorded trajectory. $L_{\rm background}$ discourages overlap with surrounding occupancy extrapolated from observations. These penalties shape candidate search, followed by explicit prefix checks before execution (\cref{sec:loop}). Their definitions and numerical limits appear in \cref{sec:implementation}.

\subsection{Context and online pressure modulation}
\label{sec:screening}\label{sec:calibration}
\looseness=1 The selected request determines the maneuver, but the pressure needed to create an encounter varies with the scene and the ego response. We therefore modulate the contact term while retaining the request's tracking objective. Recorded driving events, grouped by mechanism, actor role, scene semantics, road type, and traffic state, supply the offline reference. Let $\mu_c$ and $\mu_{p(c)}$ be the equal-weight mean over driving logs of the per-log median log-intensity in a supported context cell and in its parent (\cref{sec:knowledge_details}). The contextual multiplier is
\begin{equation}
 p_c=\operatorname{clip}\!\left[\exp\!\left(
 \frac{n_c}{n_c+30}(\mu_c-\mu_{p(c)})\right),0.75,1.35\right].
 \label{eq:core_prior}
\end{equation}
The resolver selects the most specific supported cell, $n_c$ is its event count, and unsupported contrasts use $p_c=1$. Online, observed relative position $r_t$ and velocity $u_t$ give $t_c=\operatorname{clip}(-r_t^\top u_t/\max(\|u_t\|^2,10^{-8}),0,4)$ and $d_c=\|r_t+t_cu_t\|-R$, where $R$ sums vehicle half-diagonals. With distances in meters and time in seconds,
\begin{align}
 q_t&=e^{-[d_c]_+/8}(0.5+0.5e^{-t_c/4}),\nonumber\\
 \lambda_t&=\operatorname{clip}\!\left(b_m\alpha p_c e^{\beta(1-2q_t)},0,16b_m\right).
 \label{eq:core_pressure}
\end{align}
\looseness=-1 Higher proximity $q_t$ reduces further contact pressure; $b_m$ is 20 for longitudinal requests and 2 otherwise, $\beta=1$, and $\alpha$, calibrated with the Intelligent Driver Model (IDM)~\cite{treiber2000congested}, is 0.25 on nuScenes and 4 on nuPlan. These coefficients define a bounded contact-pressure feedback schedule.

\subsection{Frozen flow generation and closed-loop execution}
\label{sec:loop}\label{sec:interpretation}
A Scene Transformer encodes the map and histories~\cite{ngiam2022scene}; a flow-matching model represents normalized acceleration/yaw-rate sequences $z$~\cite{lipman2023flow}, decoded by a unicycle model $\mathcal D$ from the target's current state $x_t^a$. At flow time $\ell$, the predicted endpoint is $\widehat z_1=z_\ell+(1-\ell)v_\theta(z_\ell,\ell\mid s_t)$. Guidance approximately solves
\begin{equation}
 \widetilde z_1^a\approx\operatorname*{arg\,min}_{\|z-\widehat z_1^a\|_\infty\leq2}
 L_t(\mathcal D(z;x_t^a),\widehat\tau_t^e;r),
 \label{eq:core_optimization}
\end{equation}
and updates the target flow by
\begin{equation}
 z_{\ell+\Delta\ell}^a=z_\ell^a+
 \Delta\ell\frac{\widetilde z_1^a-z_\ell^a}{1-\ell},\qquad\Delta\ell=1/20.
 \label{eq:core_flow}
\end{equation}
The best finite improving endpoint is retained. Sampling uses five candidates, 20 Euler steps, a 30-frame history, and a 52-frame horizon at 10\,Hz. Endpoint limited-memory BFGS (L-BFGS) runs during the final ten flow steps, with at most 30 iterations per call. On the final step, request-constructed tracking proposals may occupy existing candidate slots; up to 20 further L-BFGS iterations repair prefix feasibility within the same trust bound (\cref{sec:implementation}). These operations modify the target prediction while keeping the nuScenes-trained generator weights fixed.

The normalized-action trust region controls how far endpoint optimization departs from the model proposal. Physical checks operate on decoded trajectories: the declared request limits are speed in $[0,25]$\,m/s, longitudinal acceleration in $[-6,3]$\,m/s$^2$, lateral acceleration magnitude at most 4\,m/s$^2$, and jerk magnitude at most 10\,m/s$^3$. Candidate evaluation joins the last observed positions to the predicted prefix, linking these limits to the trajectory segment about to be executed. Optimization and prefix checks use conservative margins inside the declared limits (\cref{sec:implementation}).

\looseness=-1 Before execution, candidate checks inspect speed, acceleration, jerk, road-footprint occupancy, and target--background overlap. Finite differences across the history--prediction boundary capture discontinuities between replans. Among candidates that pass, the executor selects the lowest-cost trajectory; if none passes, it executes and records the least-violating candidate. This trajectory choice repeats at each replan, while the actor--mechanism choice remains fixed.

Every 0.5\,s, the simulator executes five frames from the ego plan and the 52-frame target prediction. New observations update the ego proposal, geometric costs, and proximity score. For a cut-in, successive replans advance the lateral reference on its original clock. Ego braking or yielding changes the encounter geometry and pressure while the lane-transition parameters remain fixed. Prefix weighting directs tracking toward the executed frames; phase gating delays contact pressure until the maneuver's active portion. The same request can therefore elicit different planner responses without changing the test specification. Training losses, initialization, and the state update appear in \cref{sec:implementation}.

\section{Evaluation}
\label{sec:exp}
\looseness=-1 We compare target contact, collateral effects, and motion distributions across four generators and five ego planners, then study request selection and mechanism-specific control on a separate cohort.
\begin{table*}[t]
\centering
\caption{\textbf{nuPlan: main comparison.} 100 shared scenes; 200 rollouts per method with learning-based planners (Diffusion Planner, Flow Planner) and 300 with rule-based planners (IDM, STRIVE, PDM-Closed). Rates are percentages; ADE/FDE are in meters. $W_1$/JSD average the acceleration and jerk discrepancies and MMD uses matched actor/time samples; $W_1(v)$, $W_1(a_{\rm lon})$, $W_1(a_{\rm lat})$, and $W_1(j)$ are the feature-wise Wasserstein distances in m/s, m/s$^2$, m/s$^2$, and m/s$^3$. Ego CR: ego contact with any vehicle; Target CR: ego contact with the initially selected target; Target--BG: target contact with non-ego vehicles; Body OR: mean fraction of target frames with more than 1\% body area off-road. Bold marks the best unrounded value per column within a planner class; the full metric set is in \cref{sec:extended}.}
\label{tab:main_nuPlan}
\begingroup\scriptsize\setlength{\tabcolsep}{2.2pt}\renewcommand{\arraystretch}{1.08}
\setlength{\tabcolsep}{0pt}\sbox{\cdtabbox}{\begin{tabular}{lrrrrrrrrrrrrr}
\toprule
 & \multicolumn{2}{c}{Contact} & \multicolumn{2}{c}{Collateral effects} & \multicolumn{3}{c}{Motion distribution} & \multicolumn{4}{c}{Feature-wise $W_1$} & \multicolumn{2}{c}{Displacement}\\
\cmidrule(lr){2-3}\cmidrule(lr){4-5}\cmidrule(lr){6-8}\cmidrule(lr){9-12}\cmidrule(l){13-14}
Method & Ego CR$\uparrow$ & Target CR$\uparrow$ & Target--BG$\downarrow$ & Body OR$\downarrow$ & $W_1\downarrow$ & JSD$\downarrow$ & MMD$\downarrow$ & $W_1(v)\downarrow$ & $W_1(a_{\rm lon})\downarrow$ & $W_1(a_{\rm lat})\downarrow$ & $W_1(j)\downarrow$ & ADE$\downarrow$ & FDE$\downarrow$\\
\midrule
\multicolumn{14}{l}{\textit{Learning-based planners (200 rollouts per method)}}\\
\rowcolor{oursrow}\textbf{CrashDojo} & \textbf{57.50} & \textbf{34.50} & \textbf{10.50} & \textbf{1.11} & \textbf{0.03799} & \textbf{0.01383} & \textbf{0.05873} & \textbf{0.3888} & \textbf{0.0305} & \textbf{0.0504} & \textbf{0.0331} & \textbf{0.750} & \textbf{1.519}\\
\addlinespace[2pt]
SAFE-SIM & 55.50 & \textbf{34.50} & 20.00 & 3.97 & 0.04138 & 0.01523 & 0.06048 & 0.4212 & 0.0327 & 0.0558 & 0.0357 & 0.791 & 1.654\\
CCDiff-DDPM & 21.50 & 7.50 & 52.00 & 11.63 & 0.06649 & 0.02590 & 0.08386 & 0.5216 & 0.0579 & 0.0971 & 0.0445 & 0.949 & 2.278\\
CCDiff-FM & 25.00 & 7.50 & 43.50 & 7.54 & 0.06280 & 0.02486 & 0.07537 & 0.5433 & 0.0507 & 0.0933 & 0.0443 & 0.922 & 2.096\\
\midrule
\multicolumn{14}{l}{\textit{Rule-based planners (300 rollouts per method)}}\\
\rowcolor{oursrow}\textbf{CrashDojo} & \textbf{59.00} & \textbf{9.67} & \textbf{18.67} & \textbf{1.79} & \textbf{0.03672} & \textbf{0.01360} & \textbf{0.06437} & \textbf{0.4620} & \textbf{0.0289} & \textbf{0.0475} & \textbf{0.0339} & 1.058 & \textbf{2.176}\\
\addlinespace[2pt]
SAFE-SIM & 54.33 & \textbf{9.67} & 26.00 & 5.22 & 0.04068 & 0.01518 & 0.06811 & 0.4829 & 0.0315 & 0.0546 & 0.0360 & \textbf{1.048} & 2.221\\
CCDiff-DDPM & 35.33 & 6.67 & 53.00 & 11.81 & 0.06630 & 0.02676 & 0.09021 & 0.5589 & 0.0599 & 0.0955 & 0.0435 & 1.194 & 2.742\\
CCDiff-FM & 33.67 & 4.00 & 43.33 & 7.69 & 0.06193 & 0.02445 & 0.08285 & 0.5858 & 0.0509 & 0.0912 & 0.0437 & 1.163 & 2.569\\
\bottomrule
\end{tabular}}%
\setlength{\tabcolsep}{\dimexpr(\textwidth-\wd\cdtabbox)/28\relax}%
\begin{tabular}{lrrrrrrrrrrrrr}
\toprule
 & \multicolumn{2}{c}{Contact} & \multicolumn{2}{c}{Collateral effects} & \multicolumn{3}{c}{Motion distribution} & \multicolumn{4}{c}{Feature-wise $W_1$} & \multicolumn{2}{c}{Displacement}\\
\cmidrule(lr){2-3}\cmidrule(lr){4-5}\cmidrule(lr){6-8}\cmidrule(lr){9-12}\cmidrule(l){13-14}
Method & Ego CR$\uparrow$ & Target CR$\uparrow$ & Target--BG$\downarrow$ & Body OR$\downarrow$ & $W_1\downarrow$ & JSD$\downarrow$ & MMD$\downarrow$ & $W_1(v)\downarrow$ & $W_1(a_{\rm lon})\downarrow$ & $W_1(a_{\rm lat})\downarrow$ & $W_1(j)\downarrow$ & ADE$\downarrow$ & FDE$\downarrow$\\
\midrule
\multicolumn{14}{l}{\textit{Learning-based planners (200 rollouts per method)}}\\
\rowcolor{oursrow}\textbf{CrashDojo} & \textbf{57.50} & \textbf{34.50} & \textbf{10.50} & \textbf{1.11} & \textbf{0.03799} & \textbf{0.01383} & \textbf{0.05873} & \textbf{0.3888} & \textbf{0.0305} & \textbf{0.0504} & \textbf{0.0331} & \textbf{0.750} & \textbf{1.519}\\
\addlinespace[2pt]
SAFE-SIM & 55.50 & \textbf{34.50} & 20.00 & 3.97 & 0.04138 & 0.01523 & 0.06048 & 0.4212 & 0.0327 & 0.0558 & 0.0357 & 0.791 & 1.654\\
CCDiff-DDPM & 21.50 & 7.50 & 52.00 & 11.63 & 0.06649 & 0.02590 & 0.08386 & 0.5216 & 0.0579 & 0.0971 & 0.0445 & 0.949 & 2.278\\
CCDiff-FM & 25.00 & 7.50 & 43.50 & 7.54 & 0.06280 & 0.02486 & 0.07537 & 0.5433 & 0.0507 & 0.0933 & 0.0443 & 0.922 & 2.096\\
\midrule
\multicolumn{14}{l}{\textit{Rule-based planners (300 rollouts per method)}}\\
\rowcolor{oursrow}\textbf{CrashDojo} & \textbf{59.00} & \textbf{9.67} & \textbf{18.67} & \textbf{1.79} & \textbf{0.03672} & \textbf{0.01360} & \textbf{0.06437} & \textbf{0.4620} & \textbf{0.0289} & \textbf{0.0475} & \textbf{0.0339} & 1.058 & \textbf{2.176}\\
\addlinespace[2pt]
SAFE-SIM & 54.33 & \textbf{9.67} & 26.00 & 5.22 & 0.04068 & 0.01518 & 0.06811 & 0.4829 & 0.0315 & 0.0546 & 0.0360 & \textbf{1.048} & 2.221\\
CCDiff-DDPM & 35.33 & 6.67 & 53.00 & 11.81 & 0.06630 & 0.02676 & 0.09021 & 0.5589 & 0.0599 & 0.0955 & 0.0435 & 1.194 & 2.742\\
CCDiff-FM & 33.67 & 4.00 & 43.33 & 7.69 & 0.06193 & 0.02445 & 0.08285 & 0.5858 & 0.0509 & 0.0912 & 0.0437 & 1.163 & 2.569\\
\bottomrule
\end{tabular}\endgroup
\end{table*}

\begin{table*}[t]
\centering
\caption{\textbf{nuScenes: main comparison.} 100 scenes; 200/300 rollouts per method with learning-based/rule-based planners. Metrics and units follow \cref{tab:main_nuPlan}.}
\label{tab:main_nuScenes}
\begingroup\scriptsize\setlength{\tabcolsep}{2.2pt}\renewcommand{\arraystretch}{1.08}
\setlength{\tabcolsep}{0pt}\sbox{\cdtabbox}{\begin{tabular}{lrrrrrrrrrrrrr}
\toprule
 & \multicolumn{2}{c}{Contact} & \multicolumn{2}{c}{Collateral effects} & \multicolumn{3}{c}{Motion distribution} & \multicolumn{4}{c}{Feature-wise $W_1$} & \multicolumn{2}{c}{Displacement}\\
\cmidrule(lr){2-3}\cmidrule(lr){4-5}\cmidrule(lr){6-8}\cmidrule(lr){9-12}\cmidrule(l){13-14}
Method & Ego CR$\uparrow$ & Target CR$\uparrow$ & Target--BG$\downarrow$ & Body OR$\downarrow$ & $W_1\downarrow$ & JSD$\downarrow$ & MMD$\downarrow$ & $W_1(v)\downarrow$ & $W_1(a_{\rm lon})\downarrow$ & $W_1(a_{\rm lat})\downarrow$ & $W_1(j)\downarrow$ & ADE$\downarrow$ & FDE$\downarrow$\\
\midrule
\multicolumn{14}{l}{\textit{Learning-based planners (200 rollouts per method)}}\\
\rowcolor{oursrow}\textbf{CrashDojo} & \textbf{32.50} & \textbf{20.50} & \textbf{12.00} & \textbf{23.68} & \textbf{0.12802} & \textbf{0.08937} & \textbf{0.11995} & 1.0880 & \textbf{0.1407} & \textbf{0.1452} & \textbf{0.0982} & 2.129 & \textbf{4.789}\\
\addlinespace[2pt]
SAFE-SIM & 23.00 & 11.50 & 16.00 & 30.54 & 0.14284 & 0.09887 & 0.12186 & \textbf{1.0167} & 0.1593 & 0.1645 & 0.1048 & \textbf{2.120} & 4.809\\
CCDiff-DDPM & 15.50 & 9.50 & 31.00 & 39.90 & 0.14999 & 0.09953 & 0.15001 & 1.4792 & 0.1772 & 0.1687 & 0.1040 & 2.935 & 7.438\\
CCDiff-FM & 14.50 & 6.50 & 30.00 & 34.96 & 0.15198 & 0.10077 & 0.13675 & 1.1718 & 0.1757 & 0.1706 & 0.1096 & 2.609 & 6.246\\
\midrule
\multicolumn{14}{l}{\textit{Rule-based planners (300 rollouts per method)}}\\
\rowcolor{oursrow}\textbf{CrashDojo} & \textbf{22.33} & 7.33 & \textbf{14.00} & \textbf{24.12} & \textbf{0.14443} & \textbf{0.09918} & \textbf{0.14027} & \textbf{1.4281} & \textbf{0.1771} & \textbf{0.1520} & \textbf{0.1042} & \textbf{2.702} & \textbf{6.601}\\
\addlinespace[2pt]
SAFE-SIM & 19.33 & 8.00 & 14.67 & 31.12 & 0.15778 & 0.10700 & 0.14328 & 1.4822 & 0.1941 & 0.1670 & 0.1122 & 2.759 & 6.768\\
CCDiff-DDPM & 13.67 & \textbf{8.67} & 33.67 & 39.58 & 0.16389 & 0.10906 & 0.16841 & 1.8399 & 0.2094 & 0.1757 & 0.1066 & 3.562 & 9.384\\
CCDiff-FM & 10.67 & 4.67 & 31.00 & 35.25 & 0.16090 & 0.10818 & 0.15427 & 1.5135 & 0.2047 & 0.1690 & 0.1089 & 3.147 & 7.958\\
\bottomrule
\end{tabular}}%
\setlength{\tabcolsep}{\dimexpr(\textwidth-\wd\cdtabbox)/28\relax}%
\begin{tabular}{lrrrrrrrrrrrrr}
\toprule
 & \multicolumn{2}{c}{Contact} & \multicolumn{2}{c}{Collateral effects} & \multicolumn{3}{c}{Motion distribution} & \multicolumn{4}{c}{Feature-wise $W_1$} & \multicolumn{2}{c}{Displacement}\\
\cmidrule(lr){2-3}\cmidrule(lr){4-5}\cmidrule(lr){6-8}\cmidrule(lr){9-12}\cmidrule(l){13-14}
Method & Ego CR$\uparrow$ & Target CR$\uparrow$ & Target--BG$\downarrow$ & Body OR$\downarrow$ & $W_1\downarrow$ & JSD$\downarrow$ & MMD$\downarrow$ & $W_1(v)\downarrow$ & $W_1(a_{\rm lon})\downarrow$ & $W_1(a_{\rm lat})\downarrow$ & $W_1(j)\downarrow$ & ADE$\downarrow$ & FDE$\downarrow$\\
\midrule
\multicolumn{14}{l}{\textit{Learning-based planners (200 rollouts per method)}}\\
\rowcolor{oursrow}\textbf{CrashDojo} & \textbf{32.50} & \textbf{20.50} & \textbf{12.00} & \textbf{23.68} & \textbf{0.12802} & \textbf{0.08937} & \textbf{0.11995} & 1.0880 & \textbf{0.1407} & \textbf{0.1452} & \textbf{0.0982} & 2.129 & \textbf{4.789}\\
\addlinespace[2pt]
SAFE-SIM & 23.00 & 11.50 & 16.00 & 30.54 & 0.14284 & 0.09887 & 0.12186 & \textbf{1.0167} & 0.1593 & 0.1645 & 0.1048 & \textbf{2.120} & 4.809\\
CCDiff-DDPM & 15.50 & 9.50 & 31.00 & 39.90 & 0.14999 & 0.09953 & 0.15001 & 1.4792 & 0.1772 & 0.1687 & 0.1040 & 2.935 & 7.438\\
CCDiff-FM & 14.50 & 6.50 & 30.00 & 34.96 & 0.15198 & 0.10077 & 0.13675 & 1.1718 & 0.1757 & 0.1706 & 0.1096 & 2.609 & 6.246\\
\midrule
\multicolumn{14}{l}{\textit{Rule-based planners (300 rollouts per method)}}\\
\rowcolor{oursrow}\textbf{CrashDojo} & \textbf{22.33} & 7.33 & \textbf{14.00} & \textbf{24.12} & \textbf{0.14443} & \textbf{0.09918} & \textbf{0.14027} & \textbf{1.4281} & \textbf{0.1771} & \textbf{0.1520} & \textbf{0.1042} & \textbf{2.702} & \textbf{6.601}\\
\addlinespace[2pt]
SAFE-SIM & 19.33 & 8.00 & 14.67 & 31.12 & 0.15778 & 0.10700 & 0.14328 & 1.4822 & 0.1941 & 0.1670 & 0.1122 & 2.759 & 6.768\\
CCDiff-DDPM & 13.67 & \textbf{8.67} & 33.67 & 39.58 & 0.16389 & 0.10906 & 0.16841 & 1.8399 & 0.2094 & 0.1757 & 0.1066 & 3.562 & 9.384\\
CCDiff-FM & 10.67 & 4.67 & 31.00 & 35.25 & 0.16090 & 0.10818 & 0.15427 & 1.5135 & 0.2047 & 0.1690 & 0.1089 & 3.147 & 7.958\\
\bottomrule
\end{tabular}\endgroup
\end{table*}

\subsection{Setup and measurements}
The main comparison uses 100 nuScenes validation scenes and 100 nuPlan-mini \texttt{evaluation} scenes~\cite{caesar2020nuscenes,caesar2021nuplan}. CrashDojo, SAFE-SIM, and two CCDiff variants, CCDiff-DDPM and CCDiff-FM (denoising-diffusion and flow-matching generators), share scenes and planners~\cite{chang2024safesim,lin2025causal}. Learning-based planners are Diffusion Planner~\cite{zheng2025diffusion} and Flow Planner~\cite{tan2025flowplanner}; rule-based planners are IDM~\cite{treiber2000congested}, our adapter of the rule-based ego planner of STRIVE~\cite{rempe2022generating}, and PDM-Closed~\cite{dauner2023parting}. This gives 200/300 rollouts per method and dataset, 4,000 in total, with each scene reused across methods and planners. Runs use seed 1 at 10\,Hz, with horizons of 10\,s on nuScenes and 7\,s on nuPlan. The main cohorts cover longitudinal, cut-in, intrusion/return, and crossing requests. All 100 nuPlan requests use native-map geometry; nuScenes has 58 native-map and 42 fallback requests. Family counts, fallback construction, and runtime details appear in \cref{sec:extended}. On nuPlan, the generator remains frozen without retraining, while a target-domain map/state adapter, behavioral prior, and guidance calibration are added.

Methods retain their implemented control scopes, including multi-actor CCDiff; uncontrolled vehicles follow the recorded log. $W_1$/JSD/MMD and ADE/FDE cover all evaluated vehicles. CrashDojo and SAFE-SIM provide the closer control-scope comparison, while CCDiff measures the behavior of joint traffic generation on the same scenes.

\label{sec:metrics}
Ego CR measures ego contact with any vehicle; Target CR measures ego contact with the target set fixed at initialization. Target--background contact (Target--BG) measures contact between that set and non-ego vehicles. Rates use completed rollouts and oriented-box contact at 10\,Hz. The target body off-road rate (Body OR) averages, within each rollout, the fraction of valid target frames with more than 1\% of body area outside the drivable region, including initial discrepancies. Native off-road rate (ORR) and all-actor average/final displacement errors (ADE/FDE) retain their original populations. ADE/FDE measure departure from logged motion. Motion comparisons use same-scene $W_1$, JSD, and paired MMD. Histogram summaries average acceleration and jerk discrepancies; speed is separate. \Cref{sec:extended} defines the reconstructed-feature discrepancy $\mathrm{RM}_{\rm rec}$ and its retained histogram support. Median relative speed at first new initial-target contact is conditioned on contact, with the corresponding counts reported alongside it.

\subsection{Main results}
\label{sec:main_results}

\Cref{tab:main_nuPlan,tab:main_nuScenes} report target contact alongside collateral effects. Relative to SAFE-SIM, CrashDojo matches or increases initial-target contact in three of four dataset--planner groups, while reducing Target--BG and Body OR in all four.

On nuPlan, CrashDojo and SAFE-SIM achieve identical Target CR under learning-based (34.50\%) and rule-based (9.67\%) planners. For the learning-based group, CrashDojo reduces Target--BG from 20.00\% to 10.50\% and Body OR from 3.97\% to 1.11\%. The rule-based group also shows lower Target--BG (18.67\% versus 26.00\%) and Body OR (1.79\% versus 5.22\%). Thus, comparable target-contact rates are achieved with fewer measured collateral effects.

On nuScenes, CrashDojo increases Target CR from 11.50\% to 20.50\% for learning-based planners, while reducing Target--BG from 16.00\% to 12.00\% and Body OR from 30.54\% to 23.68\%. For rule-based planners, Target CR is slightly lower than SAFE-SIM (7.33\% versus 8.00\%) and CCDiff-DDPM (8.67\%), but CrashDojo retains the lowest Target--BG and Body OR. Target-set sizes and native-map metric details appear in \cref{sec:extended}.

Ego CR provides a complementary measure of contact with any vehicle, rather than the specified target alone. CrashDojo has the highest group-level Ego CR, but per-planner counts (\cref{tab:per_planner_contact}) reveal heterogeneous effects: seven increases, one tie, and two decreases relative to SAFE-SIM across ten dataset--planner cells. The nuPlan rule-based aggregate, for example, combines 21 additional STRIVE contacts, seven fewer PDM-Closed contacts, and an IDM tie. These results distinguish overall collision exposure from contact with the actor selected for the challenge.

\subsection{Behavioral distributions}

\looseness=-1 These aggregate measurements cover all evaluated vehicles rather than the selected target alone. CrashDojo achieves the smallest aggregate $W_1$, JSD, and paired MMD values in all four dataset--planner groups. For learning-based planners on nuScenes, aggregate $W_1$ decreases from SAFE-SIM's 0.14284 to 0.12802, alongside the higher target-contact rate reported above. Feature-wise, CrashDojo achieves the lowest longitudinal-acceleration, lateral-acceleration, and jerk discrepancies in every group. Complete motion breakdowns and valid counts appear in \cref{tab:full_nuPlan_learning,tab:full_nuPlan_rule,tab:full_nuScenes_learning,tab:full_nuScenes_rule}.

\begin{figure*}[t]
\centering
\includegraphics[width=\textwidth]{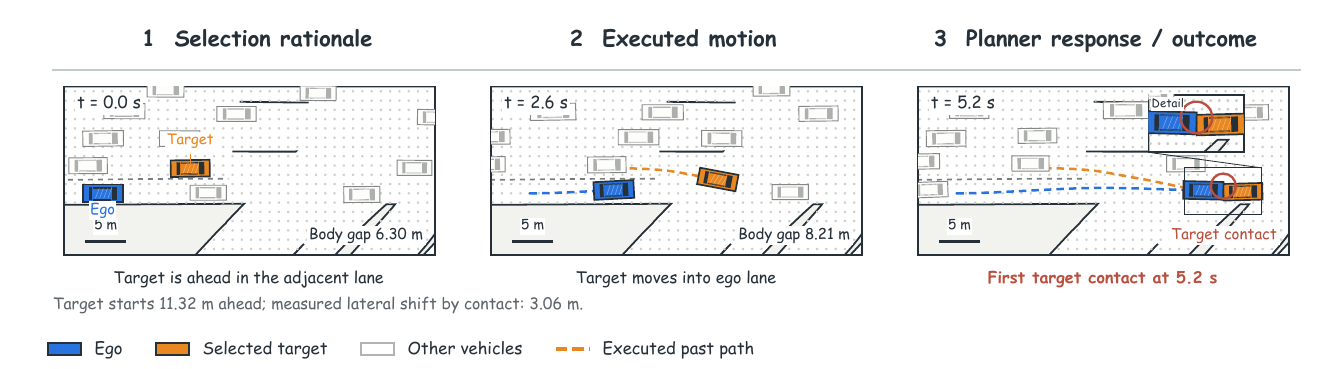}
\caption{\textbf{An executed cut-in with Flow Planner.} The selected target starts ahead in the adjacent lane, moves into the ego lane, and reaches first ego--target contact at 5.2\,s. The measured lateral shift by contact is 3.06\,m. Blue/orange denote ego/target; dashed lines show executed motion, dots show the native road mask, and red circles mark oriented-box contact. The fixed-view sequence is an example from the main-comparison cohort; Supplementary~\ref{sec:cases} provides the complete five-mechanism case set.}
\label{fig:executed_cutin}
\end{figure*}

CrashDojo also has the lowest FDE in all four groups, indicating less final displacement from logged motion over the evaluated vehicle population. \Cref{fig:executed_cutin} illustrates an individual cut-in request: the target progresses into the receiving lane, with first target contact at 5.2\,s. The body gap increases from 6.30\,m at initialization to 8.21\,m at 2.6\,s before contact. Additional measured sequences and auxiliary VLM descriptions appear in \cref{sec:cases,sec:vlm}.

\subsection{Component ablations}
\label{sec:ablation}
\begin{table*}[t]
\centering
\caption{\textbf{Component study on a 100-scene development cohort.} nuScenes, 7\,s, three seeds, $n=300$ per planner cell, with one Full baseline and actor-aligned random streams per scene--planner--seed group. CR: designated-target contact; MS: mean relative speed at first target contact over contact rollouts; $\mathrm{RM}_{\rm tgt}$: target acceleration/jerk histogram distance to a fixed real reference, distinct from supplementary $\mathrm{RM}_{\rm rec}$. Adv.ADE/Adv.FDE measure target displacement from its logged trajectory. Values are reported mean $\pm$ sample standard deviation; bold marks column optima. A4 has no adversarial guidance; -- means displacement was not reported, not undefined.}
\label{tab:ablation_summary}\label{tab:design_ablation}\label{tab:overall_guidance}
\begingroup\scriptsize\setlength{\tabcolsep}{4pt}\renewcommand{\arraystretch}{1.0}
\setlength{\tabcolsep}{0pt}\sbox{\cdtabbox}{\begin{tabular}{lrrrrr}
\toprule
Variant & CR $\uparrow$ (\%) & MS $\uparrow$ (m/s) & $\mathrm{RM}_{\rm tgt}\downarrow$ & Adv.ADE $\downarrow$ (m) & Adv.FDE $\downarrow$ (m)\\
\midrule
\multicolumn{6}{l}{\textit{IDM}}\\
\rowcolor{oursrow}\textbf{A0 Full CrashDojo} & \textbf{14.7$\pm$2.9} & \textbf{8.31$\pm$0.43} & \textbf{0.7827$\pm$0.0045} & \textbf{7.59$\pm$0.02} & \textbf{19.72$\pm$0.06}\\
A1 Random actor, canonical request & 12.3$\pm$1.5 & 3.18$\pm$0.21 & 0.8202$\pm$0.0052 & 7.96$\pm$0.12 & 20.75$\pm$0.33\\
A2 TTC actor, canonical request & 13.3$\pm$0.6 & 3.45$\pm$0.09 & 0.7898$\pm$0.0057 & 7.85$\pm$0.09 & 19.98$\pm$0.17\\
A3 Generic contact pressure & 14.0$\pm$3.6 & 4.55$\pm$0.48 & 0.8935$\pm$0.0115 & 8.44$\pm$0.11 & 22.19$\pm$0.44\\
A4 No adversarial guidance & 4.3$\pm$2.1 & 3.50$\pm$0.49 & 0.8372$\pm$0.0040 & -- & --\\
\midrule
\multicolumn{6}{l}{\textit{Diffusion Planner}}\\
\rowcolor{oursrow}\textbf{A0 Full CrashDojo} & \textbf{23.0$\pm$1.0} & \textbf{5.26$\pm$0.89} & \textbf{0.7857$\pm$0.0112} & \textbf{7.52$\pm$0.05} & \textbf{19.66$\pm$0.06}\\
A1 Random actor, canonical request & 7.7$\pm$2.1 & 4.37$\pm$0.22 & 0.8269$\pm$0.0030 & 7.88$\pm$0.10 & 20.73$\pm$0.39\\
A2 TTC actor, canonical request & 11.0$\pm$1.0 & 4.28$\pm$0.14 & 0.7862$\pm$0.0073 & 7.75$\pm$0.05 & 19.82$\pm$0.18\\
A3 Generic contact pressure & 17.3$\pm$2.1 & 4.04$\pm$0.63 & 0.8814$\pm$0.0083 & 8.06$\pm$0.25 & 22.05$\pm$0.58\\
A4 No adversarial guidance & 9.0$\pm$1.0 & 3.31$\pm$0.49 & 0.8272$\pm$0.0147 & -- & --\\
\bottomrule
\end{tabular}}%
\setlength{\tabcolsep}{\dimexpr(\textwidth-\wd\cdtabbox)/12\relax}%
\begin{tabular}{lrrrrr}
\toprule
Variant & CR $\uparrow$ (\%) & MS $\uparrow$ (m/s) & $\mathrm{RM}_{\rm tgt}\downarrow$ & Adv.ADE $\downarrow$ (m) & Adv.FDE $\downarrow$ (m)\\
\midrule
\multicolumn{6}{l}{\textit{IDM}}\\
\rowcolor{oursrow}\textbf{A0 Full CrashDojo} & \textbf{14.7$\pm$2.9} & \textbf{8.31$\pm$0.43} & \textbf{0.7827$\pm$0.0045} & \textbf{7.59$\pm$0.02} & \textbf{19.72$\pm$0.06}\\
A1 Random actor, canonical request & 12.3$\pm$1.5 & 3.18$\pm$0.21 & 0.8202$\pm$0.0052 & 7.96$\pm$0.12 & 20.75$\pm$0.33\\
A2 TTC actor, canonical request & 13.3$\pm$0.6 & 3.45$\pm$0.09 & 0.7898$\pm$0.0057 & 7.85$\pm$0.09 & 19.98$\pm$0.17\\
A3 Generic contact pressure & 14.0$\pm$3.6 & 4.55$\pm$0.48 & 0.8935$\pm$0.0115 & 8.44$\pm$0.11 & 22.19$\pm$0.44\\
A4 No adversarial guidance & 4.3$\pm$2.1 & 3.50$\pm$0.49 & 0.8372$\pm$0.0040 & -- & --\\
\midrule
\multicolumn{6}{l}{\textit{Diffusion Planner}}\\
\rowcolor{oursrow}\textbf{A0 Full CrashDojo} & \textbf{23.0$\pm$1.0} & \textbf{5.26$\pm$0.89} & \textbf{0.7857$\pm$0.0112} & \textbf{7.52$\pm$0.05} & \textbf{19.66$\pm$0.06}\\
A1 Random actor, canonical request & 7.7$\pm$2.1 & 4.37$\pm$0.22 & 0.8269$\pm$0.0030 & 7.88$\pm$0.10 & 20.73$\pm$0.39\\
A2 TTC actor, canonical request & 11.0$\pm$1.0 & 4.28$\pm$0.14 & 0.7862$\pm$0.0073 & 7.75$\pm$0.05 & 19.82$\pm$0.18\\
A3 Generic contact pressure & 17.3$\pm$2.1 & 4.04$\pm$0.63 & 0.8814$\pm$0.0083 & 8.06$\pm$0.25 & 22.05$\pm$0.58\\
A4 No adversarial guidance & 9.0$\pm$1.0 & 3.31$\pm$0.49 & 0.8272$\pm$0.0147 & -- & --\\
\bottomrule
\end{tabular}\endgroup
\end{table*}

The development cohort contains 100 separate nuScenes scenes with at least two eligible actors, ordered by scene attributes fixed before any rollout, with native candidates only. The reported study has five conditions, two planners, a 7\,s horizon, and three seeds (3,000 rollouts). CR is designated-target contact. $\mathrm{RM}_{\rm tgt}$ averages histogram $W_1$ for signed target body-frame accelerations and longitudinal jerk against a frozen real-reference pool (\cref{sec:target_motion_metric}); its features and population differ from $\mathrm{RM}_{\rm rec}$.

A1 draws uniformly over eligible actor IDs; A2 picks the earliest constant-velocity TTC, breaking ties by actor ID. Both retain candidate enumeration and choose a within-actor request by canonical family/parameter/geometry identity, independently of the evidence ordering in \cref{eq:core_selection}. They therefore compare complete actor--request selection policies (\cref{sec:impl_selection}), rather than isolating the evidence score. For Diffusion Planner, mean CR falls from Full's 23.0\% to 7.7\% for A1 and 11.0\% for A2. The corresponding IDM values are 14.7\%, 12.3\%, and 13.3\%. A2's target-motion discrepancy is also closer to Full's than A1's for both planners.

\enlargethispage{\baselineskip}%
A3 replaces mechanism-specific guidance with generic contact pressure. It disables action tracking, mechanism-phase contact gating, crossing/arrival objectives, and mechanism-conditioned proposal/completion logic, retaining common regularizers and contextual pressure. Relative to Full, mean $\mathrm{RM}_{\rm tgt}$ increases by 14.2\% with IDM and 12.2\% with Diffusion Planner, while Adv.ADE and Adv.FDE also increase for both planners. Mean CR is 14.0\% versus Full's 14.7\% under IDM and 17.3\% versus 23.0\% under Diffusion Planner. Thus, the mechanism-specific bundle improves target-motion measurements for both planners, while its contact-rate benefit is more pronounced for Diffusion Planner than for IDM.

\looseness=-1 The reported means and standard deviations further imply that Full's lower $\mathrm{RM}_{\rm tgt}$, Adv.ADE, and Adv.FDE hold across all three reported seeds, even allowing for display rounding (\cref{sec:seed_consistency}). Full's CR advantage over A3 is likewise consistent across the three Diffusion Planner seeds. For IDM, only the motion improvement is seed-consistent.

Removing adversarial guidance (A4; zero tracking and contact-pressure weights) reduces mean CR from 14.7\% to 4.3\% with IDM and from 23.0\% to 9.0\% with Diffusion Planner. The remaining contacts have lower mean relative speeds (MS), while target-motion discrepancy is higher than Full's for both planners. Together, A1--A4 demonstrate the effects of request selection and mechanism-specific control on target contact and motion. 

\enlargethispage{\baselineskip}%
\section{Conclusion}
\label{sec:conclusion}

\looseness=-1 CrashDojo connects accident-grounded interaction requests to persistent request-guided control for closed-loop planner testing. The request preserves the selected actor, maneuver parameters, and timing while target motion adapts to the ego response. Across two datasets and five planners, CrashDojo matches or exceeds SAFE-SIM in initial-target contact in three of four groups, while reducing target--background contact and target body off-road rates in all four groups. These results demonstrate a practical connection between explicit challenge construction and adaptive closed-loop execution.

{
    \footnotesize
    \bibliographystyle{ieeenat_fullname}
    \bibliography{main}
}

\appendix
\label{sec:supplementary_start}
\makeatletter\setlength{\@fptop}{0pt}\makeatother
\renewcommand{\dblfloatpagefraction}{0.5}\renewcommand{\floatpagefraction}{0.5}
\section{Detailed formulation and implementation}
\label{sec:implementation}
This section supplies the detailed references, penalties, contextual factors, and sampling procedure for the evaluated configuration in \cref{sec:method}. All main-comparison tables retain that configuration.

\subsection{Accident-grounded actor--mechanism selection}
\label{sec:impl_knowledge}\label{sec:impl_selection}
A nearby actor need not be able to initiate the requested interaction. We therefore select an actor jointly with a mechanism and its parameters. A road--accident lookup informed by pre-crash typologies~\cite{najm2007precrash} associates mechanisms with road layouts and participant roles. For example, longitudinal braking requires a relevant same-corridor relationship, a cut-in requires an adjacent receiving lane, and crossing requires intersecting routes. Let
\begin{equation}
 \mathcal C(s_0)=\{\xi=(a,m,\eta):\xi\text{ is enumerated for }s_0\}
 \label{eq:method_candidates}
\end{equation}
be the candidate pool. Enumeration is distinct from qualification: each candidate receives seven checks covering traffic situation, road structure, accident support, vehicle capability/history, nominal spatiotemporal conflict, maneuver intensity, and configuration completeness.

\paragraph{Evidence-ordered selection.}
Let $N_{\rm fail}(\xi)$ and $N_{\rm unk}(\xi)$ count failed and insufficient-evidence checks. The implemented selector minimizes the following tuple in lexicographic order:
\begin{align}
 \xi^\star&=\operatorname*{arg\,min}_{\xi\in\mathcal C(s_0)}^{\rm lex}\mathbf k(\xi),\label{eq:method_selection}\\
 \mathbf k(\xi)&=\bigl(N_{\rm fail},N_{\rm unk},R_{\rm acc},R_{\rm conflict},\nonumber\\[-2pt]
 &\hspace{15mm}R_{\rm intensity},\bar d_{\min},\bar A_{\rm lon},\mathrm{ID}(a)\bigr).\nonumber
\end{align}
Here $R_{\rm acc}$ prioritizes mechanism statistics supported by similar cases over statistics alone and weaker evidence; $R_{\rm conflict}$ orders nominal contact before near-miss proximity before more distant encounters. $R_{\rm intensity}$ orders the declared motion-intensity tiers. The final geometric terms are the minimum body gap rounded to $10^{-3}$\,m and the sum of peak braking and acceleration magnitudes rounded to $10^{-2}$\,m/s$^2$; actor ID breaks remaining ties. Thus weak evidence cannot be compensated by an arbitrary weighted collision score. The nominal conflict terms are computed during offline preparation and do not become future-state inputs to online guidance.

Selection produces the frozen request and its evidence record, not a learned collision probability or a validity certificate. In particular, a selected tuple may still have failed or unknown checks. The main comparison includes requests with different geometry provenance, documented in \cref{sec:extended}. Empty or unsupported requests and runtime failures remain identifiable in the request ledger.

\paragraph{Actor-ablation request selection.}
For A1/A2, each eligible actor is first assigned a canonical request using the tuple of family name, sorted-key parameter serialization, and sorted-key geometry serialization. Geometry includes the route, corridor, intervention window, offset bounds, and template/progress fields; evidence verdicts and the ordering scores of \cref{eq:core_selection} are excluded. A1 samples uniformly over eligible actor IDs with the declared seed, and A2 selects minimum constant-velocity TTC with actor-ID ties. Choosing Full's actor does not restore Full's differently ranked request.

\subsection{Geometry-specific intervention objectives}
\label{sec:impl_guidance}
Given the selected request, guidance targets its specified maneuver and a challenging encounter. At time $t$, consider a target prediction of $H$ steps with $\Delta=0.1$\,s and absolute times $T_k=t+k\Delta$. Project target positions onto the stored route to obtain station $s_k$ and signed lateral offset $d_k$; let $v_k$ denote speed. References are evaluated on the persistent request clock, so replanning does not restart the reference schedule.

\paragraph{Mechanism references.}
For a smooth transition, define $S(u)=10\bar u^3-15\bar u^4+6\bar u^5$, where $\bar u=\operatorname{clip}(u,0,1)$. The basic speed and cut-in references are
\begin{align}
 v_k^\star&=v_0+(v_{\rm goal}-v_0)S\!\left(\frac{T_k-t_v}{T_v}\right),\nonumber\\
 d_k^\star&=d_0+\Delta d\,S\!\left(\frac{T_k-t_0}{T_{\rm lc}}\right).
 \label{eq:references}
\end{align}
Longitudinal requests retain $d_k^\star=d_0$. Intrusion-and-return requests use
\begin{align}
 d_k^\star={}&d_0+A S\!\left(\frac{T_k-t_0}{T_{\rm in}}\right)\nonumber\\
 &+(d_{\rm ret}-d_0-A)S\!\left(\frac{T_k-t_0-T_{\rm in}-T_{\rm hold}}{T_{\rm out}}\right),
 \label{eq:method_intrusion}
\end{align}
where $A$ is the requested lateral increment and $d_{\rm ret}$ the return offset. When boundary geometry is available, the intrusion reference is constructed relative to that boundary and penalties inspect the oriented vehicle footprint, distinguishing partial intrusion from full lane entry. Request-specific phase-completion references additionally govern settling and return behavior.

The reference family also supports bounded-jerk speed transitions. For $\vartheta=\operatorname{clip}(T-t_v,0,T_v)$, ramp duration $T_r$, acceleration magnitude $a_\star$, and direction $\sigma=\operatorname{sign}(v_{\rm goal}-v_0)$, this template is
\begin{equation}
 v^\star(T)=v_0+\frac{\sigma a_\star}{2T_r}
 \left(\vartheta^2-[\vartheta-T_r]_+^2-[\vartheta-T_v+T_r]_+^2\right).
 \label{eq:method_speed_jerk}
\end{equation}
The frozen request specifies the profile and any phase composition; a smooth reference alone does not guarantee that generated motion satisfies the limits.

\paragraph{Tracking and encounter timing.}
Define the per-step tracking error and the executed-prefix emphasis as
\begin{align}
 E_k&=\left(\frac{d_k-d_k^\star}{\sigma_d}\right)^2+
       \left(\frac{v_k-v_k^\star}{\sigma_v}\right)^2,\nonumber\\
 L_{\rm track}&=\frac1H\sum_{k=1}^H E_k+
                  \frac3h\sum_{k=1}^{h}E_k,\qquad h=\min(5,H),
 \label{eq:method_tracking}
\end{align}
with $\sigma_d=0.35$\,m and $\sigma_v=0.5$\,m/s in every evaluated request. Prefix weighting and mechanism-specific completion penalties discourage repeatedly postponing the action into an unexecuted prediction tail. We write
\begin{equation}
 L_{\rm action}=L_{\rm track}+L_{\rm phase}+L_{\rm boundary}+L_{\rm arrival},
 \label{eq:method_action}
\end{equation}
where terms are enabled only for the applicable mechanism and supported geometry. $L_{\rm phase}$ penalizes incomplete speed/lateral transitions; $L_{\rm boundary}$ checks intrusion depth and return against the shared boundary.

For crossing and merging, $s_k^\star=s_{\rm init}+\int_0^{T_k}v^\star(u)\,du$ specifies route progress from the initial route station $s_{\rm init}$ on the request clock. With frozen arrival time $T_\star$, arrival station $s_\star$, and prediction index $k_\star=\operatorname{round}((T_\star-t)/\Delta)$, the enabled arrival terms take the form
\begin{equation}
 L_{\rm arrival}=\frac1H\sum_k\left(\frac{s_k-s_k^\star}{2\,\mathrm m}\right)^2+
 \mathbf1_{\{k_\star\in[1,H]\}}\left(\frac{s_{k_\star}-s_\star}{5\,\mathrm m}\right)^2.
 \label{eq:method_arrival}
\end{equation}
The progress term is enabled by the request. This construction guides the target toward a specified encounter location and time; the realized timing still depends on closed-loop execution and the ego response.

\paragraph{Contact pressure.}
Let $\mathcal B_k^a$ and $\mathcal B_k^e$ be the oriented target and ego rectangles. For their four separating axes $\mathcal N_k$, define the signed separation proxy
\begin{equation}
 D_k=\max_{n\in\mathcal N_k}
 \left(|n^\top(p_k^a-p_k^e)|-h_k^a(n)-h_k^e(n)\right),
 \label{eq:method_sat}
\end{equation}
where $h_k^i(n)$ is the projected half-extent. Positive values indicate a separating axis; nonpositive values indicate overlap or contact. Over the mechanism-active prediction indices $\mathcal I_t$ within $\mathcal W$, contact search minimizes
\begin{equation}
 L_{\rm contact}=\min_{k\in\mathcal I_t}
 \left\{\frac{[D_k+\epsilon]_+^2}{4}+\rho(T_k-t)\right\},
 \label{eq:contact}
\end{equation}
with $\epsilon=0.05$\,m and $\rho=1$ in the stored cost scaling; the term is zero when the active set is empty. The delay cost favors earlier attainable contact. For cut-ins, $\mathcal I_t$ retains times $T_k\geq t_0+T_{\rm lc}/2$ within $\mathcal W$; intrusion masks select their corresponding phase. When crossing geometry and active indices are available, the pressure term additionally includes $0.1L_{\rm cross}$, where
\begin{equation}
 L_{\rm cross}=\min_{k\in\mathcal I_t}
 \left([\|p_k^a-p_\times\|-R]_+^2+[\|p_k^e-p_\times\|-R]_+^2\right).
 \label{eq:method_crossing}
\end{equation}
Here $p_\times$ is the stored crossing point and $R$ sums the vehicles' half-diagonals. This body-scale timing proxy encourages simultaneous proximity to the conflict region; executed contact is evaluated separately from this objective.

\paragraph{Behavioral and feasibility penalties.}
Denote the pressure term, including the crossing augmentation when applicable, by $L_{\rm pressure}$. The target objective is
\begin{equation}
 \begin{split}
 L_t={}&w_mL_{\rm action}+\lambda_tL_{\rm pressure}
       +0.1L_{\rm prior}+30L_{\rm limits}\\
      &+L_{\rm deviation}+100L_{\rm background},
 \end{split}
 \label{eq:guidecost}
\end{equation}
where $w_m=5$ for longitudinal requests and 20 otherwise. $L_{\rm prior}$ is the mean squared change in decoded acceleration and yaw rate relative to the endpoint supplied before the current optimization call. It is a motion regularizer, distinct from the contextual prior used to set $\lambda_t$.
Writing $(\bar a_k,\bar\omega_k)$ for those pre-optimization actions gives
\begin{equation}
 L_{\rm prior}=\frac{1}{2H}\sum_{k=1}^H
 \bigl((a_k-\bar a_k)^2+(\omega_k-\bar\omega_k)^2\bigr).
 \label{eq:method_motion_prior}
\end{equation}

Limit penalties use the squared band violation
\begin{equation}
 B(x;l,u,s)=\frac{[l-x]_+^2+[x-u]_+^2}{s^2}.
 \label{eq:method_band}
\end{equation}
$L_{\rm limits}$ combines speed, longitudinal/lateral acceleration, jerk, route-offset, and footprint penalties. The declared limits are $v\in[0,25]$\,m/s, $a_{\rm lon}\in[-6,3]$\,m/s$^2$, $|a_{\rm lat}|\leq4$\,m/s$^2$, and jerk magnitude at most 10\,m/s$^3$. Additional finite-difference terms join observed history to the predicted prefix to expose discontinuities across replans; conservative optimization margins lie inside the acceptance limits.

$L_{\rm deviation}$ penalizes excess over soft ADE/FDE budgets of 2/4\,m relative to the first pre-guidance predicted endpoint saved within the replan. For anchor positions $\bar p_k$, define $\bar e=H^{-1}\sum_k\|p_k-\bar p_k\|$ and $e_H=\|p_H-\bar p_H\|$. With the request's fixed deviation weights,
\begin{equation}
 L_{\rm deviation}=w_{\rm ADE}[\bar e-2\,\mathrm m]_+^2+
 w_{\rm FDE}[e_H-4\,\mathrm m]_+^2.
 \label{eq:method_deviation}
\end{equation}
This is a model-generated anchor, not the logged target future. $L_{\rm background}$ penalizes predicted overlap with non-target occupancy extrapolated from observations with bounded uncertainty. Specifically, for uncertainty-adjusted target--background separations $\widetilde D_k^{a,j}$ and request margin $\epsilon_{\rm bg}$, let $b_k=\max_{j\notin\{e,a\}}[\epsilon_{\rm bg}-\widetilde D_k^{a,j}]_+^2$. The avoidance cost is
\begin{equation}
 L_{\rm background}=\frac1H\sum_k b_k+\max_k b_k+\frac4h\sum_{k=1}^{h}b_k,
 \label{eq:method_background}
\end{equation}
with zero cost when no observed background actor is available. The mean, peak, and prefix terms penalize persistent, severe, and immediately executable overlap, respectively. These are optimization penalties and candidate checks, not hard guarantees on the final rollout.

\subsection{Guidance-strength implementation}
\label{sec:impl_screening}\label{sec:impl_calibration}
A fixed contact coefficient applies the same pressure to contexts with different motion intensities. CrashDojo therefore combines an offline behavioral reference with an online proximity signal. The former is fixed at scene initialization; the latter follows the evolving ego--target interaction. Both modulate the contact-pressure term in \cref{eq:guidecost}, while mechanism tracking retains its own weight.

\paragraph{Hierarchical behavioral reference.}
Reference driving events are grouped by dataset, mechanism, actor role, scene semantics, road type, and traffic state. Let $\mu_c$ be the stored log-balanced summary of a supported context cell, and $\mu_{p(c)}$ that of its mechanism/role parent. The summary takes the median log motion intensity within each source log and then averages these medians across logs. Only reference rows marked \texttt{fit}, meaning they passed the derivative-window and observation checks of \cref{sec:knowledge_details}, are used; event-intensity definitions and the calibration procedure are given in \cref{sec:knowledge_details}. The implemented factor is
\begin{equation}
 p_c=\operatorname{clip}\!\left[
 \exp\!\left(\frac{n_c}{n_c+\kappa}(\mu_c-\mu_{p(c)})\right),p_{\min},p_{\max}\right],
 \label{eq:prior}
\end{equation}
where $n_c$ is the event count, $\kappa=30$, and $(p_{\min},p_{\max})=(0.75,1.35)$. Shrinkage reduces the influence of sparsely estimated differences, and clipping bounds their effect on pressure.

The resolver searches from joint context to semantic/road, semantic, road, and family/role groups. Fine cells require at least 20 events, 10 scenes, and three logs; broader cells require 10 events, five scenes, and three logs. If role-specific context is sparse, explicitly pooled-role cells can supply a reference; the chosen parent and support counts are retained. If only the parent is available, the difference vanishes; unsupported families use $p_c=1$. Thus the method does not assume that every requested mechanism has a supported fine-grained conditional distribution.

\paragraph{Online risk adjustment.}
Let $r_t=p_t^a-p_t^e$ and $u_t=\dot p_t^a-\dot p_t^e$, estimated from current poses and observed histories. A bounded constant-velocity closest-approach proxy gives
\begin{align}
 t_c&=\operatorname{clip}\!\left(-\frac{r_t^\top u_t}{\max(\|u_t\|^2,10^{-8})},0,4\right),\nonumber\\
 d_c&=\|r_t+t_cu_t\|-R,\nonumber\\
 q_t&=\exp\!\left(-\frac{[d_c]_+}{8}\right)
       \left(\frac12+\frac12e^{-t_c/4}\right).
 \label{eq:method_risk}
\end{align}
Distances and times are in meters and seconds. $q_t\in(0,1]$ is a proximity score, not a calibrated collision probability or the box-based contact test in \cref{eq:method_sat}. The pressure coefficient is
\begin{equation}
 \lambda_t=\operatorname{clip}\!\left(b_m\alpha p_c e^{\beta(1-2q_t)},0,16b_m\right).
 \label{eq:calibration}
\end{equation}
For fixed context and away from clipping, $\partial\lambda_t/\partial q_t=-2\beta\lambda_t$: increasing proximity reduces further contact pressure, while a weak encounter receives more pressure. This is a control heuristic for the objective, not a claim that realized risk varies monotonically with $\lambda_t$.

The main-experiment configuration uses $b_m=20$ for longitudinal requests and 2 otherwise, $\beta=1$, and frozen IDM-calibrated domain scales $\alpha=0.25$ on nuScenes and 4 on nuPlan. Consequently, transfer uses a frozen generator with target-domain calibration. The mapping from a cost coefficient to realized acceleration, gap, or arrival time depends on the model and constraints; $\lambda_t$ is not a behavior quantile. No explicit conditional-distribution loss is enabled in these requests, so behavioral agreement is assessed on executed motion.

\subsection{Frozen flow generation and closed-loop execution}
\label{sec:impl_loop}\label{sec:impl_interpretation}
The guidance operates on candidate actions rather than model parameters. A Scene Transformer encodes map and motion histories~\cite{ngiam2022scene}; the flow model represents normalized acceleration/yaw-rate sequences $z$~\cite{lipman2023flow}. A unicycle decoder $\mathcal D(z;x_t^a)$ produces target positions, headings, speeds, and physical actions. During training, $z_\ell=(1-\ell)z_0+\ell z_1$ interpolates Gaussian and recorded actions. The network predicts $v_\theta(z_\ell,\ell\mid s_t)$, and the configured reconstruction objective is applied to the decoded predicted endpoint $z_\ell+(1-\ell)v_\theta$. Training collision/off-road auxiliary weights are zero. The nuScenes-trained checkpoint remains frozen for all evaluated interventions.

\paragraph{Bounded endpoint optimization.}
At flow time $\ell<1$, let
\begin{equation}
 \widehat z_1=z_\ell+(1-\ell)v_\theta(z_\ell,\ell\mid s_t).
 \label{eq:method_endpoint}
\end{equation}
For the selected actor, guidance approximately solves
\begin{equation}
 \widetilde z_1^a\approx
 \operatorname*{arg\,min}_{\|z-\widehat z_1^a\|_\infty\leq\delta}
 L_t\!\left(\mathcal D(z;x_t^a),\widehat\tau_t^e;r\right),
 \label{eq:method_optimization}
\end{equation}
where $\delta=2$ in normalized action units. Non-target endpoints are not optimized by this target objective. The solver retains the best finite improving endpoint found, rather than assuming that the nonconvex optimum is reached. The target flow update becomes
\begin{equation}
 z_{\ell+\Delta\ell}^a=z_\ell^a+
 \Delta\ell\frac{\widetilde z_1^a-z_\ell^a}{1-\ell},\qquad\Delta\ell=1/20.
 \label{eq:flow}
\end{equation}
The implementation draws five candidates using 20 Euler steps, a 30-frame history, and a 52-frame prediction horizon. L-BFGS is applied during the final ten flow steps with at most 30 iterations per call. On the final step, request-dependent tracking proposals can seed existing candidate slots; up to 20 further L-BFGS iterations repair prefix feasibility within the same trust bound before execution. These operations do not update the generator weights or smooth an already executed trajectory.

\paragraph{Receding-horizon interaction.}
Let $\widehat\tau_t^{a,\star}$ be the candidate selected with feasibility priority, and let $\tau^{j,\mathrm{bg}}$ denote surrounding-vehicle motion. With $h=5$ frames, the simulator advances according to
\begin{equation}
 s_{t+h\Delta}=\mathcal F_h\!\left(s_t,
 \widehat\tau_t^e[1{:}h],\widehat\tau_t^{a,\star}[1{:}h],
 \{\tau^{j,\mathrm{bg}}_{t:t+h\Delta}\}_{j\notin\{e,a\}}\right).
 \label{eq:method_closed_loop}
\end{equation}
Here $\mathcal F_h$ denotes the simulator and history update, not a new learned dynamics model. Every 0.5\,s, the ego proposal, geometric costs, and proximity score are recomputed from the updated scene. Target identity, request timing, and the offline contextual factor persist. This allows the target to respond to ego braking or yielding while preserving the intended maneuver clock.

If no candidate passes the checks, the executor uses the least-violating candidate, and the rollout is scored like any other; its violations are not excluded. This execution rule is unrelated to actor selection. The main comparison rolls out 70 frames (7\,s) on nuPlan and 100 frames (10\,s) on nuScenes. The resulting system examines planner responses to the selected target's evolving challenge within the surrounding traffic.

\section{Accident knowledge and calibration details}
\label{sec:knowledge_details}
\paragraph{From accident records to maneuver requests.}
The lookup combines the Najm pre-crash typology, a road-conditioned Crash Report Sampling System (CRSS) frequency table, and coded Crash Investigation Sampling System (CISS) participant records~\cite{najm2007precrash,nhtsa_crss,nhtsa_ciss}. Accident records supply type frequencies, participant-role codes, and available pre-crash motion evidence. The association of these records with simulator road topology, candidate roles, and parameterized maneuvers is an explicit engineering mapping. CISS context matches are code-based retrievals, not manually verified reconstructions of individual crashes. The archived extraction inferred junction/control code groupings from cross-tabulations with crash types; these groupings are retained as implementation assumptions rather than treated as authoritative case interpretations.

\begin{table*}[t]
\centering
\caption{\textbf{Examples of the frozen road--role--mechanism mapping.} Geometry determines applicability; accident support supplies an additional ordering criterion. Signal state is used only when observed. Najm $n$ denotes pre-crash scenario $n$ of~\cite{najm2007precrash}.}
\label{tab:knowledge_mapping}
\begingroup\small\setlength{\tabcolsep}{5pt}\renewcommand{\arraystretch}{1.18}
\begin{tabular}{@{}>{\raggedright\arraybackslash}p{.24\textwidth}>{\raggedright\arraybackslash}p{.20\textwidth}>{\raggedright\arraybackslash}p{.235\textwidth}>{\raggedright\arraybackslash}p{.255\textwidth}@{}}
\toprule
Road relation & Candidate role & Accident mapping & Request geometry\\
\midrule
Same directed corridor & Relevant leader & Lead-vehicle deceleration; Najm 25 & C1: longitudinal speed reduction\\
Adjacent same-direction lanes & Adjacent-lane actor & Lane change; Najm 18 & C2: transition across the shared boundary\\
Adjacent same-direction lanes & Adjacent-lane actor & Same-direction drifting; Najm 19 & C3A/C3B: intrusion and specified return\\
Intersecting routes & Crossing or turning actor & Crossing-path or turn-across-path type, resolved from headings and control & C4: route progress and conflict-point arrival\\
Distinct incoming routes, shared downstream lane & Merging actor & Same-direction merge outside a junction; turn-into-path at a junction & C5: progress into the common downstream corridor\\
\bottomrule
\end{tabular}\endgroup
\end{table*}

Mechanisms are coded C1 (lead braking), C2 (cut-in), C3A/C3B (intrusion without/with specified return), C4 (crossing), and C5 (merge); \cref{tab:knowledge_mapping} lists the road--role--mechanism mapping. For example, the frozen C2 request for \texttt{scene-0559} identifies same-direction adjacent lanes with a shared native-map boundary. It specifies a lateral increment of 2.813\,m starting at 1.4\,s over a 3.5\,s transition, together with a speed reference from 2.373 to 4.373\,m/s. These are request parameters, not measured outcomes. The target ID, reference route, boundary, and mechanism clock remain fixed; current pair geometry and the ego proposal update at each replan.

\paragraph{Seven recorded selection checks.}
Each candidate has a pass, fail, or insufficient-evidence verdict for each check. A failure takes precedence over missing evidence when combining subchecks. The selector orders these verdicts as in \cref{eq:method_selection}; selection does not imply seven passes.
\begin{enumerate}
\item \textit{Traffic situation.} C1 requires a target ahead of the ego and within 2\,m laterally of its path. Adjacent-lane mechanisms require a lateral separation of 1.2--6.5\,m and a compatible longitudinal encounter. Missing signal state remains unknown rather than being inferred from a collision.
\item \textit{Road structure.} C1 checks the relevant directed corridor and intervening vehicles. C2/C3 check same-direction adjacency and a shared boundary. C4 requires a crossing angle of at least $30^\circ$ and an ego path within 2\,m of the crossing point; C5 checks distinct incoming routes and an ego path within 2\,m of their common downstream entry.
\item \textit{Accident support.} A mapped type must account for at least 0.5\% of the stored road-conditioned CRSS distribution. At least three context-matched CISS records give the stronger evidence level; otherwise supported type statistics remain a separate level. Retrieval filters junction/control context and speed limits up to 72\,km/h when recorded.
\item \textit{Vehicle capability and history.} The actor must exist in the raw tracks, have a complete preceding 1\,s history, be non-parked and moving at least 0.5\,m/s, match a driving lane, and have length 3--13\,m and width 1.4--3\,m. A discrepancy above 2\,m/s between stored and displacement-derived speed fails this check.
\item \textit{Nominal spatiotemporal conflict.} The nominal maneuver is compared with the recorded ego continuation within the intervention window. Contact or a body gap no larger than 2\,m supplies the encounter evidence. Earlier contact, contact with another actor before ego contact, and missing ego frames receive separate verdicts.
\item \textit{Maneuver intensity.} Nominal braking/acceleration is compared first with the stored road-class reference, then with the corresponding CISS pre-crash 90th-percentile reference when available. Engineering limits additionally constrain speed, acceleration, lateral acceleration, jerk, and corridor coverage.
\item \textit{Configuration completeness.} Target, mechanism, guidance function, parameters, weights, start, and intervention window must be present. Initial body-overlap area above 0.01\,m$^2$ fails the check; the stored 0.02\,s timing diagnostic flags sampling-sensitive encounters.
\end{enumerate}

\paragraph{Observed motion intensity.}
The behavioral prior uses recorded driving events, separately from accident-record strength tiers. Positions are interpolated to a 10\,Hz grid, and an 11-sample quadratic Savitzky--Golay fit estimates acceleration. Only event samples with the full centered 1\,s derivative window and adequate source observations are retained. For an event $E$,
\begin{equation}
 I(E)=\begin{cases}
 \max_{k\in E}[-a_{\parallel,k}]_+, & \mathrm{C1},\\
 \max_{k\in E}|a_{\perp,k}|, & \mathrm{C2,C3A,C3B},\\
 (\operatorname{PET}(E)+0.5\,\mathrm{s})^{-1}, & \mathrm{C4},\\
 \max_{k\in E}|a_{\parallel,k}|, & \mathrm{C5}.
 \end{cases}
\end{equation}
For C4, the post-encroachment time (PET) is the nonnegative gap between verified conflict-zone occupancy intervals; overlapping intervals give zero. Both actors require one contiguous occupancy episode and at least 0.5\,s of clear observation on either side. Intensities are compared within a mechanism family, so their different units are not mixed into one event score. Jerk enters the feasibility checks and output metrics, not this prior-intensity definition.

For context cell $c$, the implementation takes a median log intensity within each source log, then gives logs equal weight:
\begin{equation}
 \mu_c=\frac{1}{|\mathcal L_c|}\sum_{\ell\in\mathcal L_c}
 \operatorname{median}_{E\in(c,\ell)}\log I(E).
\end{equation}

\paragraph{Freezing the global pressure scale.}
Calibration uses 12 scenes per dataset, the IDM planner, and seed 1, comparing $(\alpha,\beta)\in\{(0.25,0),(4,0),(0.25,1),(1,1),(4,1)\}$ and a fixed-reference setting. The frozen ordering maximizes contacts passing the archived validity checks, then minimizes invalid execution prefixes and target--background contacts. Ties prefer $\alpha$ closer to one, lower $\beta$, then the stored condition ID. The selected coefficients, $(0.25,1)$ for nuScenes and $(4,1)$ for nuPlan, are inherited by the main comparison without refitting on its results.

\section{Executed mechanism examples}
\label{sec:cases}
\begin{figure*}[tp]
\centering
\includegraphics[width=\textwidth,height=.86\textheight,keepaspectratio]{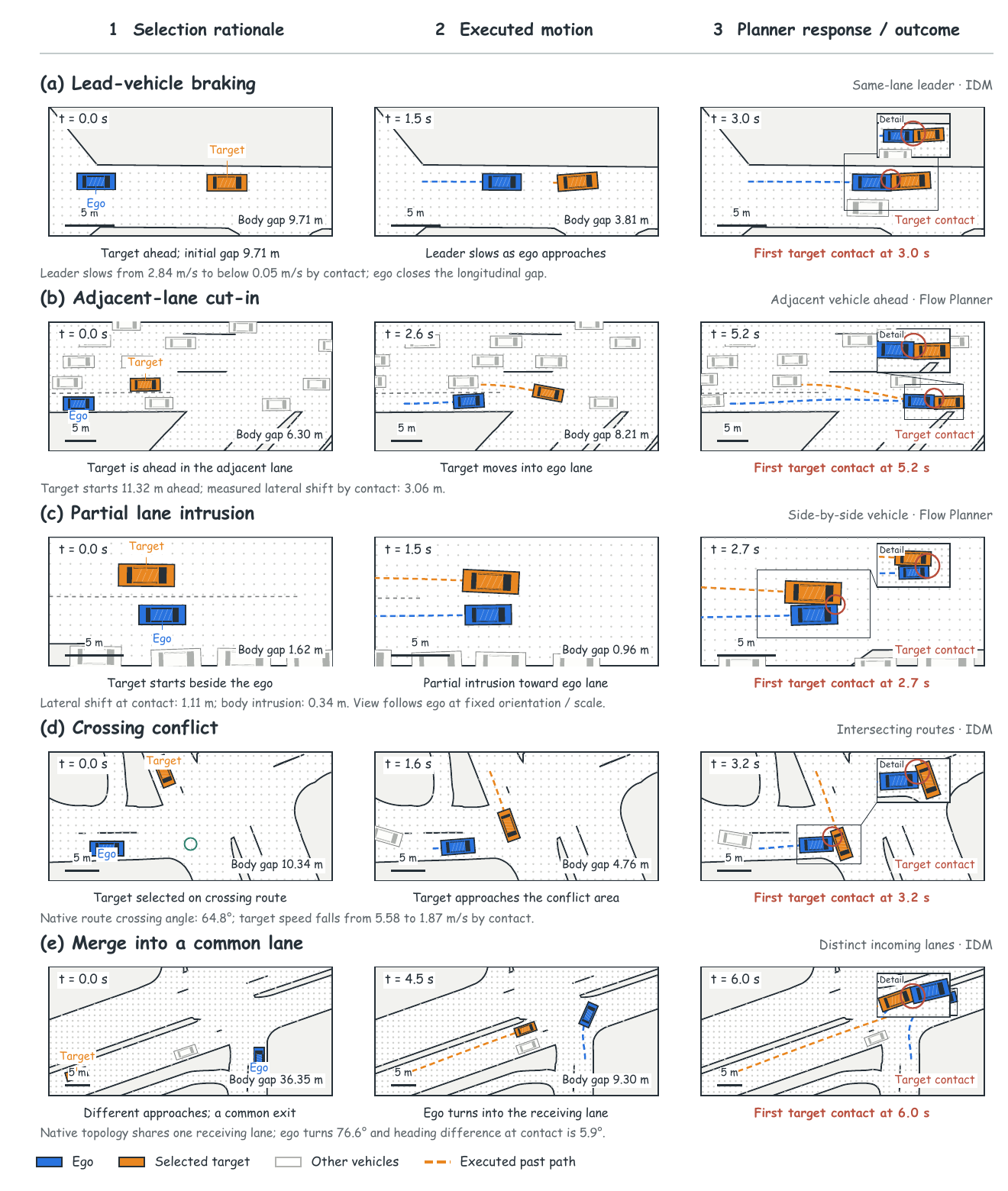}
\caption{\textbf{Measured ego--target contacts across five mechanisms.} Rows show (a) lead braking, (b) adjacent-lane cut-in, (c) partial lane intrusion, (d) crossing, and (e) merging into a common lane. The last column of each row shows the first target contact, checked by oriented-box overlap at 10\,Hz. Blue/orange mark ego/target; gray marks other vehicles. Dots show native road masks, dashed paths show executed motion, and red circles mark contact. The green circle in (d) marks the map-derived route intersection. Views are fixed within (a,b,d,e); (c) translates with the ego at fixed orientation and scale. Rows (a--d) come from the main-comparison cohorts; (e) is an additional archived merge case. These selected examples illustrate realized motion and contact geometry.}
\label{fig:qualitative}
\end{figure*}

\Cref{fig:qualitative} follows five executed interactions from the initial actor--road relationship through the maneuver to first ego--target contact. The cases cover lead-vehicle braking, adjacent-lane cut-in, partial lane intrusion, crossing, and merging into a common lane, with first contacts at 3.0, 5.2, 2.7, 3.2, and 6.0\,s, respectively. The cut-in target starts ahead and moves laterally by 3.06\,m before contact; the partial-intrusion target moves laterally by 1.11\,m, with 0.34\,m of body intrusion. In the merge case, distinct incoming lanes share a receiving lane, and the ego turn brings the vehicles into near-aligned contact. Geometry and timestamps are taken from archived trajectories.

The first four examples come from the completed main-comparison cohorts; the merge example is an additional archived rollout. They illustrate how selected roles and requested mechanisms relate to executed motion. Aggregate challenge and behavior are assessed by the main tables.

These five clips, seven additional cases, and a 36-case stratified extension are also checked by the VLM plausibility screen of \cref{sec:vlm}.

\section{Extended results and metric protocol}
\label{sec:extended}
\label{sec:protocol_details}
\Cref{tab:full_nuPlan_learning,tab:full_nuPlan_rule,tab:full_nuScenes_learning,tab:full_nuScenes_rule} report 25 metrics from the completed main comparison on the two 100-scene cohorts, split into learning-based and rule-based ego-planner groups. Each method has 200 completed rollouts in the learning-based group and 300 in the rule-based group of each dataset. Values and ranks are generated from the verified group aggregates, retaining raw precision before display rounding. N/A is not replaced by zero. Methods share scenes and planners within each table, but native controlled-agent scopes and valid frame counts differ. A percentage rounded to 0.000 can still be nonzero.
\begin{table*}[p]
\centering
\caption{\textbf{nuPlan: 25 metrics for learning-based ego planners.} Diffusion Planner and Flow Planner; the fixed 100-scene cohort has 200 rollouts per method. Rates are percentages; CDC-road is a fraction. Motion-distribution scores follow \cref{sec:extended}. Second lines retain counts/denominators or conditional $n$; rollout means use $n=200$ unless shown otherwise. $\mathrm{RM}_{\rm rec}$ uses a group-specific reference pool and the retained-feature definition in \cref{sec:extended}. Heterogeneous-hardware time is unranked; other ranks use unrounded values and shared ranks for ties.}
\label{tab:full_nuPlan_learning}
\begingroup\small\setlength{\tabcolsep}{4pt}\renewcommand{\arraystretch}{1.12}
\begin{tabular*}{\textwidth}{@{\extracolsep{\fill}}lrrrrr@{}}
\toprule
Metric & \textbf{CrashDojo} & SAFE-SIM & CCDiff-DDPM & CCDiff-FM & Our rank\\
\midrule
Ego collision $\uparrow$ & \makecell[r]{\textbf{57.50}\\[-1pt]{\footnotesize\color{black!65} 115/200}} & \makecell[r]{55.50\\[-1pt]{\footnotesize\color{black!65} 111/200}} & \makecell[r]{21.50\\[-1pt]{\footnotesize\color{black!65} 43/200}} & \makecell[r]{25.00\\[-1pt]{\footnotesize\color{black!65} 50/200}} & 1\\
Initial-target contact $\uparrow$ & \makecell[r]{\textbf{34.50}\\[-1pt]{\footnotesize\color{black!65} 69/200}} & \makecell[r]{\textbf{34.50}\\[-1pt]{\footnotesize\color{black!65} 69/200}} & \makecell[r]{7.50\\[-1pt]{\footnotesize\color{black!65} 15/200}} & \makecell[r]{7.50\\[-1pt]{\footnotesize\color{black!65} 15/200}} & 1\\
Target--background $\downarrow$ & \makecell[r]{\textbf{10.50}\\[-1pt]{\footnotesize\color{black!65} 21/200}} & \makecell[r]{20.00\\[-1pt]{\footnotesize\color{black!65} 40/200}} & \makecell[r]{52.00\\[-1pt]{\footnotesize\color{black!65} 104/200}} & \makecell[r]{43.50\\[-1pt]{\footnotesize\color{black!65} 87/200}} & 1\\
ADE (m) $\downarrow$ & \textbf{0.750} & 0.791 & 0.949 & 0.922 & 1\\
FDE (m) $\downarrow$ & \textbf{1.519} & 1.654 & 2.278 & 2.096 & 1\\
Native ORR $\downarrow$ & 0.045 & \textbf{0.002} & 0.445 & 0.030 & 3\\
TTC at $-0.5$ s (s) $\downarrow$ & \makecell[r]{\textbf{0.444}\\[-1pt]{\footnotesize\color{black!65} $n=68$}} & \makecell[r]{0.455\\[-1pt]{\footnotesize\color{black!65} $n=67$}} & \makecell[r]{0.536\\[-1pt]{\footnotesize\color{black!65} $n=15$}} & \makecell[r]{0.479\\[-1pt]{\footnotesize\color{black!65} $n=14$}} & 1\\
\addlinespace[4pt]
Other-pair CR $\downarrow$ & \makecell[r]{\textbf{52.00}\\[-1pt]{\footnotesize\color{black!65} 104/200}} & \makecell[r]{55.00\\[-1pt]{\footnotesize\color{black!65} 110/200}} & \makecell[r]{66.50\\[-1pt]{\footnotesize\color{black!65} 133/200}} & \makecell[r]{62.00\\[-1pt]{\footnotesize\color{black!65} 124/200}} & 1\\
Controlled--background CR $\downarrow$ & \makecell[r]{\textbf{10.50}\\[-1pt]{\footnotesize\color{black!65} 21/200}} & \makecell[r]{20.00\\[-1pt]{\footnotesize\color{black!65} 40/200}} & \makecell[r]{34.00\\[-1pt]{\footnotesize\color{black!65} 68/200}} & \makecell[r]{30.00\\[-1pt]{\footnotesize\color{black!65} 60/200}} & 1\\
Initial-target center off-road $\downarrow$ & \makecell[r]{\textbf{0.000}\\[-1pt]{\footnotesize\color{black!65} 0/14,000}} & \makecell[r]{0.014\\[-1pt]{\footnotesize\color{black!65} 2/14,000}} & \makecell[r]{4.391\\[-1pt]{\footnotesize\color{black!65} 2,846/64,820}} & \makecell[r]{2.200\\[-1pt]{\footnotesize\color{black!65} 1,426/64,820}} & 1\\
Body off-road, rollout mean $\downarrow$ & \textbf{1.11} & 3.97 & 11.63 & 7.54 & 1\\
\addlinespace[4pt]
First-contact relative speed (m/s) $\uparrow$ & \makecell[r]{3.565\\[-1pt]{\footnotesize\color{black!65} $n=69$}} & \makecell[r]{\textbf{3.763}\\[-1pt]{\footnotesize\color{black!65} $n=69$}} & \makecell[r]{1.735\\[-1pt]{\footnotesize\color{black!65} $n=15$}} & \makecell[r]{1.821\\[-1pt]{\footnotesize\color{black!65} $n=15$}} & 2\\
Reconstructed $\mathrm{RM}_{\rm rec}\downarrow$ & 0.8158 & \textbf{0.7884} & 1.1848 & 1.1446 & 2\\
Mean request time (s) $\downarrow$ & 467.5 & 239.2 & 492.9 & 246.4 & --\\
\addlinespace[4pt]
Mean body area outside $\downarrow$ & \textbf{0.031} & 0.336 & 4.756 & 2.447 & 1\\
Any frame with $>1\%$ outside $\downarrow$ & \makecell[r]{\textbf{4.00}\\[-1pt]{\footnotesize\color{black!65} 8/200}} & \makecell[r]{15.50\\[-1pt]{\footnotesize\color{black!65} 31/200}} & \makecell[r]{74.50\\[-1pt]{\footnotesize\color{black!65} 149/200}} & \makecell[r]{60.00\\[-1pt]{\footnotesize\color{black!65} 120/200}} & 1\\
Body off-road, pooled frames $\downarrow$ & \makecell[r]{\textbf{1.11}\\[-1pt]{\footnotesize\color{black!65} 156/14,000}} & \makecell[r]{3.97\\[-1pt]{\footnotesize\color{black!65} 556/14,000}} & \makecell[r]{11.61\\[-1pt]{\footnotesize\color{black!65} 7,523/64,820}} & \makecell[r]{7.57\\[-1pt]{\footnotesize\color{black!65} 4,909/64,820}} & 1\\
\addlinespace[4pt]
Wasserstein $\downarrow$ & \makecell[r]{\textbf{0.03799}\\[-1pt]{\footnotesize\color{black!65} $n=600$}} & \makecell[r]{0.04138\\[-1pt]{\footnotesize\color{black!65} $n=600$}} & \makecell[r]{0.06649\\[-1pt]{\footnotesize\color{black!65} $n=600$}} & \makecell[r]{0.06280\\[-1pt]{\footnotesize\color{black!65} $n=600$}} & 1\\
JSD (nats) $\downarrow$ & \makecell[r]{\textbf{0.01383}\\[-1pt]{\footnotesize\color{black!65} $n=600$}} & \makecell[r]{0.01523\\[-1pt]{\footnotesize\color{black!65} $n=600$}} & \makecell[r]{0.02590\\[-1pt]{\footnotesize\color{black!65} $n=600$}} & \makecell[r]{0.02486\\[-1pt]{\footnotesize\color{black!65} $n=600$}} & 1\\
MMD $\downarrow$ & \textbf{0.05873} & 0.06048 & 0.08386 & 0.07537 & 1\\
CDC-road $\uparrow$ & \makecell[r]{\textbf{1.00000}\\[-1pt]{\footnotesize\color{black!65} $n=18$}} & \makecell[r]{\textbf{1.00000}\\[-1pt]{\footnotesize\color{black!65} $n=18$}} & \makecell[r]{0.97750\\[-1pt]{\footnotesize\color{black!65} $n=18$}} & \makecell[r]{0.98278\\[-1pt]{\footnotesize\color{black!65} $n=18$}} & 1\\
\addlinespace[4pt]
$W_1$(speed), m/s $\downarrow$ & \textbf{0.3888} & 0.4212 & 0.5216 & 0.5433 & 1\\
$W_1(a_{\rm lon})$, m/s$^2$ $\downarrow$ & \textbf{0.0305} & 0.0327 & 0.0579 & 0.0507 & 1\\
$W_1(a_{\rm lat})$, m/s$^2$ $\downarrow$ & \textbf{0.0504} & 0.0558 & 0.0971 & 0.0933 & 1\\
$W_1$(jerk), m/s$^3$ $\downarrow$ & \textbf{0.0331} & 0.0357 & 0.0445 & 0.0443 & 1\\
\bottomrule
\end{tabular*}\endgroup
\end{table*}

\begin{table*}[p]
\centering
\caption{\textbf{nuPlan: 25 metrics for rule-based ego planners.} IDM, STRIVE and PDM-Closed; the fixed 100-scene cohort has 300 rollouts per method. Rates are percentages; CDC-road is a fraction. Motion-distribution scores follow \cref{sec:extended}. Second lines retain counts/denominators or conditional $n$; rollout means use $n=300$ unless shown otherwise. $\mathrm{RM}_{\rm rec}$ uses a group-specific reference pool and the retained-feature definition in \cref{sec:extended}. Heterogeneous-hardware time is unranked; other ranks use unrounded values and shared ranks for ties.}
\label{tab:full_nuPlan_rule}
\begingroup\small\setlength{\tabcolsep}{4pt}\renewcommand{\arraystretch}{1.12}
\begin{tabular*}{\textwidth}{@{\extracolsep{\fill}}lrrrrr@{}}
\toprule
Metric & \textbf{CrashDojo} & SAFE-SIM & CCDiff-DDPM & CCDiff-FM & Our rank\\
\midrule
Ego collision $\uparrow$ & \makecell[r]{\textbf{59.00}\\[-1pt]{\footnotesize\color{black!65} 177/300}} & \makecell[r]{54.33\\[-1pt]{\footnotesize\color{black!65} 163/300}} & \makecell[r]{35.33\\[-1pt]{\footnotesize\color{black!65} 106/300}} & \makecell[r]{33.67\\[-1pt]{\footnotesize\color{black!65} 101/300}} & 1\\
Initial-target contact $\uparrow$ & \makecell[r]{\textbf{9.67}\\[-1pt]{\footnotesize\color{black!65} 29/300}} & \makecell[r]{\textbf{9.67}\\[-1pt]{\footnotesize\color{black!65} 29/300}} & \makecell[r]{6.67\\[-1pt]{\footnotesize\color{black!65} 20/300}} & \makecell[r]{4.00\\[-1pt]{\footnotesize\color{black!65} 12/300}} & 1\\
Target--background $\downarrow$ & \makecell[r]{\textbf{18.67}\\[-1pt]{\footnotesize\color{black!65} 56/300}} & \makecell[r]{26.00\\[-1pt]{\footnotesize\color{black!65} 78/300}} & \makecell[r]{53.00\\[-1pt]{\footnotesize\color{black!65} 159/300}} & \makecell[r]{43.33\\[-1pt]{\footnotesize\color{black!65} 130/300}} & 1\\
ADE (m) $\downarrow$ & 1.058 & \textbf{1.048} & 1.194 & 1.163 & 2\\
FDE (m) $\downarrow$ & \textbf{2.176} & 2.221 & 2.742 & 2.569 & 1\\
Native ORR $\downarrow$ & 0.018 & \textbf{0.004} & 0.440 & 0.041 & 2\\
TTC at $-0.5$ s (s) $\downarrow$ & \makecell[r]{\textbf{0.398}\\[-1pt]{\footnotesize\color{black!65} $n=29$}} & \makecell[r]{0.428\\[-1pt]{\footnotesize\color{black!65} $n=29$}} & \makecell[r]{0.443\\[-1pt]{\footnotesize\color{black!65} $n=20$}} & \makecell[r]{0.426\\[-1pt]{\footnotesize\color{black!65} $n=11$}} & 1\\
\addlinespace[4pt]
Other-pair CR $\downarrow$ & \makecell[r]{66.67\\[-1pt]{\footnotesize\color{black!65} 200/300}} & \makecell[r]{67.00\\[-1pt]{\footnotesize\color{black!65} 201/300}} & \makecell[r]{71.33\\[-1pt]{\footnotesize\color{black!65} 214/300}} & \makecell[r]{\textbf{65.67}\\[-1pt]{\footnotesize\color{black!65} 197/300}} & 2\\
Controlled--background CR $\downarrow$ & \makecell[r]{\textbf{18.67}\\[-1pt]{\footnotesize\color{black!65} 56/300}} & \makecell[r]{26.00\\[-1pt]{\footnotesize\color{black!65} 78/300}} & \makecell[r]{34.33\\[-1pt]{\footnotesize\color{black!65} 103/300}} & \makecell[r]{30.33\\[-1pt]{\footnotesize\color{black!65} 91/300}} & 1\\
Initial-target center off-road $\downarrow$ & \makecell[r]{\textbf{0.000}\\[-1pt]{\footnotesize\color{black!65} 0/21,000}} & \makecell[r]{0.257\\[-1pt]{\footnotesize\color{black!65} 54/21,000}} & \makecell[r]{4.359\\[-1pt]{\footnotesize\color{black!65} 4,238/97,230}} & \makecell[r]{2.229\\[-1pt]{\footnotesize\color{black!65} 2,167/97,230}} & 1\\
Body off-road, rollout mean $\downarrow$ & \textbf{1.79} & 5.22 & 11.81 & 7.69 & 1\\
\addlinespace[4pt]
First-contact relative speed (m/s) $\uparrow$ & \makecell[r]{2.082\\[-1pt]{\footnotesize\color{black!65} $n=29$}} & \makecell[r]{2.217\\[-1pt]{\footnotesize\color{black!65} $n=29$}} & \makecell[r]{4.074\\[-1pt]{\footnotesize\color{black!65} $n=20$}} & \makecell[r]{\textbf{4.282}\\[-1pt]{\footnotesize\color{black!65} $n=12$}} & 4\\
Reconstructed $\mathrm{RM}_{\rm rec}\downarrow$ & 0.8479 & \textbf{0.8142} & 1.2066 & 1.1699 & 2\\
Mean request time (s) $\downarrow$ & 474.4 & 241.4 & 497.3 & 250.3 & --\\
\addlinespace[4pt]
Mean body area outside $\downarrow$ & \textbf{0.068} & 0.536 & 4.736 & 2.501 & 1\\
Any frame with $>1\%$ outside $\downarrow$ & \makecell[r]{\textbf{5.00}\\[-1pt]{\footnotesize\color{black!65} 15/300}} & \makecell[r]{18.67\\[-1pt]{\footnotesize\color{black!65} 56/300}} & \makecell[r]{76.00\\[-1pt]{\footnotesize\color{black!65} 228/300}} & \makecell[r]{60.00\\[-1pt]{\footnotesize\color{black!65} 180/300}} & 1\\
Body off-road, pooled frames $\downarrow$ & \makecell[r]{\textbf{1.79}\\[-1pt]{\footnotesize\color{black!65} 376/21,000}} & \makecell[r]{5.22\\[-1pt]{\footnotesize\color{black!65} 1,097/21,000}} & \makecell[r]{11.78\\[-1pt]{\footnotesize\color{black!65} 11,451/97,230}} & \makecell[r]{7.72\\[-1pt]{\footnotesize\color{black!65} 7,511/97,230}} & 1\\
\addlinespace[4pt]
Wasserstein $\downarrow$ & \makecell[r]{\textbf{0.03672}\\[-1pt]{\footnotesize\color{black!65} $n=900$}} & \makecell[r]{0.04068\\[-1pt]{\footnotesize\color{black!65} $n=900$}} & \makecell[r]{0.06630\\[-1pt]{\footnotesize\color{black!65} $n=900$}} & \makecell[r]{0.06193\\[-1pt]{\footnotesize\color{black!65} $n=900$}} & 1\\
JSD (nats) $\downarrow$ & \makecell[r]{\textbf{0.01360}\\[-1pt]{\footnotesize\color{black!65} $n=900$}} & \makecell[r]{0.01518\\[-1pt]{\footnotesize\color{black!65} $n=900$}} & \makecell[r]{0.02676\\[-1pt]{\footnotesize\color{black!65} $n=900$}} & \makecell[r]{0.02445\\[-1pt]{\footnotesize\color{black!65} $n=900$}} & 1\\
MMD $\downarrow$ & \textbf{0.06437} & 0.06811 & 0.09021 & 0.08285 & 1\\
CDC-road $\uparrow$ & \makecell[r]{\textbf{1.00000}\\[-1pt]{\footnotesize\color{black!65} $n=27$}} & \makecell[r]{\textbf{1.00000}\\[-1pt]{\footnotesize\color{black!65} $n=27$}} & \makecell[r]{0.97813\\[-1pt]{\footnotesize\color{black!65} $n=27$}} & \makecell[r]{0.98350\\[-1pt]{\footnotesize\color{black!65} $n=27$}} & 1\\
\addlinespace[4pt]
$W_1$(speed), m/s $\downarrow$ & \textbf{0.4620} & 0.4829 & 0.5589 & 0.5858 & 1\\
$W_1(a_{\rm lon})$, m/s$^2$ $\downarrow$ & \textbf{0.0289} & 0.0315 & 0.0599 & 0.0509 & 1\\
$W_1(a_{\rm lat})$, m/s$^2$ $\downarrow$ & \textbf{0.0475} & 0.0546 & 0.0955 & 0.0912 & 1\\
$W_1$(jerk), m/s$^3$ $\downarrow$ & \textbf{0.0339} & 0.0360 & 0.0435 & 0.0437 & 1\\
\bottomrule
\end{tabular*}\endgroup
\end{table*}

\begin{table*}[p]
\centering
\caption{\textbf{nuScenes: 25 metrics for learning-based ego planners.} Diffusion Planner and Flow Planner; the fixed 100-scene cohort has 200 rollouts per method. Rates are percentages; CDC-road is a fraction. Motion-distribution scores follow \cref{sec:extended}. Second lines retain counts/denominators or conditional $n$; rollout means use $n=200$ unless shown otherwise. $\mathrm{RM}_{\rm rec}$ uses a group-specific reference pool and the retained-feature definition in \cref{sec:extended}. Heterogeneous-hardware time is unranked; other ranks use unrounded values and shared ranks for ties.}
\label{tab:full_nuScenes_learning}
\begingroup\small\setlength{\tabcolsep}{4pt}\renewcommand{\arraystretch}{1.12}
\begin{tabular*}{\textwidth}{@{\extracolsep{\fill}}lrrrrr@{}}
\toprule
Metric & \textbf{CrashDojo} & SAFE-SIM & CCDiff-DDPM & CCDiff-FM & Our rank\\
\midrule
Ego collision $\uparrow$ & \makecell[r]{\textbf{32.50}\\[-1pt]{\footnotesize\color{black!65} 65/200}} & \makecell[r]{23.00\\[-1pt]{\footnotesize\color{black!65} 46/200}} & \makecell[r]{15.50\\[-1pt]{\footnotesize\color{black!65} 31/200}} & \makecell[r]{14.50\\[-1pt]{\footnotesize\color{black!65} 29/200}} & 1\\
Initial-target contact $\uparrow$ & \makecell[r]{\textbf{20.50}\\[-1pt]{\footnotesize\color{black!65} 41/200}} & \makecell[r]{11.50\\[-1pt]{\footnotesize\color{black!65} 23/200}} & \makecell[r]{9.50\\[-1pt]{\footnotesize\color{black!65} 19/200}} & \makecell[r]{6.50\\[-1pt]{\footnotesize\color{black!65} 13/200}} & 1\\
Target--background $\downarrow$ & \makecell[r]{\textbf{12.00}\\[-1pt]{\footnotesize\color{black!65} 24/200}} & \makecell[r]{16.00\\[-1pt]{\footnotesize\color{black!65} 32/200}} & \makecell[r]{31.00\\[-1pt]{\footnotesize\color{black!65} 62/200}} & \makecell[r]{30.00\\[-1pt]{\footnotesize\color{black!65} 60/200}} & 1\\
ADE (m) $\downarrow$ & 2.129 & \textbf{2.120} & 2.935 & 2.609 & 2\\
FDE (m) $\downarrow$ & \textbf{4.789} & 4.809 & 7.438 & 6.246 & 1\\
Native ORR $\downarrow$ & 3.015 & \textbf{2.560} & 6.189 & 4.344 & 2\\
TTC at $-0.5$ s (s) $\downarrow$ & \makecell[r]{\textbf{0.454}\\[-1pt]{\footnotesize\color{black!65} $n=41$}} & \makecell[r]{0.471\\[-1pt]{\footnotesize\color{black!65} $n=22$}} & \makecell[r]{0.470\\[-1pt]{\footnotesize\color{black!65} $n=18$}} & \makecell[r]{0.471\\[-1pt]{\footnotesize\color{black!65} $n=12$}} & 1\\
\addlinespace[4pt]
Other-pair CR $\downarrow$ & \makecell[r]{\textbf{31.00}\\[-1pt]{\footnotesize\color{black!65} 62/200}} & \makecell[r]{32.50\\[-1pt]{\footnotesize\color{black!65} 65/200}} & \makecell[r]{34.50\\[-1pt]{\footnotesize\color{black!65} 69/200}} & \makecell[r]{36.50\\[-1pt]{\footnotesize\color{black!65} 73/200}} & 1\\
Controlled--background CR $\downarrow$ & \makecell[r]{\textbf{12.00}\\[-1pt]{\footnotesize\color{black!65} 24/200}} & \makecell[r]{16.00\\[-1pt]{\footnotesize\color{black!65} 32/200}} & \makecell[r]{21.00\\[-1pt]{\footnotesize\color{black!65} 42/200}} & \makecell[r]{16.50\\[-1pt]{\footnotesize\color{black!65} 33/200}} & 1\\
Initial-target center off-road $\downarrow$ & \makecell[r]{\textbf{18.040}\\[-1pt]{\footnotesize\color{black!65} 3,608/20,000}} & \makecell[r]{21.765\\[-1pt]{\footnotesize\color{black!65} 4,353/20,000}} & \makecell[r]{29.576\\[-1pt]{\footnotesize\color{black!65} 22,655/76,600}} & \makecell[r]{26.069\\[-1pt]{\footnotesize\color{black!65} 19,969/76,600}} & 1\\
Body off-road, rollout mean $\downarrow$ & \textbf{23.68} & 30.54 & 39.90 & 34.96 & 1\\
\addlinespace[4pt]
First-contact relative speed (m/s) $\uparrow$ & \makecell[r]{3.854\\[-1pt]{\footnotesize\color{black!65} $n=41$}} & \makecell[r]{\textbf{4.313}\\[-1pt]{\footnotesize\color{black!65} $n=23$}} & \makecell[r]{3.265\\[-1pt]{\footnotesize\color{black!65} $n=19$}} & \makecell[r]{4.019\\[-1pt]{\footnotesize\color{black!65} $n=13$}} & 3\\
Reconstructed $\mathrm{RM}_{\rm rec}\downarrow$ & \textbf{0.2051} & 0.2459 & 0.2712 & 0.2736 & 1\\
Mean request time (s) $\downarrow$ & 353.3 & 218.4 & 299.3 & 180.2 & --\\
\addlinespace[4pt]
Mean body area outside $\downarrow$ & \textbf{18.397} & 22.613 & 31.738 & 27.935 & 1\\
Any frame with $>1\%$ outside $\downarrow$ & \makecell[r]{\textbf{32.00}\\[-1pt]{\footnotesize\color{black!65} 64/200}} & \makecell[r]{39.00\\[-1pt]{\footnotesize\color{black!65} 78/200}} & \makecell[r]{88.50\\[-1pt]{\footnotesize\color{black!65} 177/200}} & \makecell[r]{77.00\\[-1pt]{\footnotesize\color{black!65} 154/200}} & 1\\
Body off-road, pooled frames $\downarrow$ & \makecell[r]{\textbf{23.67}\\[-1pt]{\footnotesize\color{black!65} 4,735/20,000}} & \makecell[r]{30.54\\[-1pt]{\footnotesize\color{black!65} 6,108/20,000}} & \makecell[r]{37.79\\[-1pt]{\footnotesize\color{black!65} 28,948/76,600}} & \makecell[r]{33.12\\[-1pt]{\footnotesize\color{black!65} 25,371/76,600}} & 1\\
\addlinespace[4pt]
Wasserstein $\downarrow$ & \makecell[r]{\textbf{0.12802}\\[-1pt]{\footnotesize\color{black!65} $n=594$}} & \makecell[r]{0.14284\\[-1pt]{\footnotesize\color{black!65} $n=594$}} & \makecell[r]{0.14999\\[-1pt]{\footnotesize\color{black!65} $n=594$}} & \makecell[r]{0.15198\\[-1pt]{\footnotesize\color{black!65} $n=594$}} & 1\\
JSD (nats) $\downarrow$ & \makecell[r]{\textbf{0.08937}\\[-1pt]{\footnotesize\color{black!65} $n=594$}} & \makecell[r]{0.09887\\[-1pt]{\footnotesize\color{black!65} $n=594$}} & \makecell[r]{0.09953\\[-1pt]{\footnotesize\color{black!65} $n=594$}} & \makecell[r]{0.10077\\[-1pt]{\footnotesize\color{black!65} $n=594$}} & 1\\
MMD $\downarrow$ & \textbf{0.11995} & 0.12186 & 0.15001 & 0.13675 & 1\\
CDC-road $\uparrow$ & \makecell[r]{\textbf{0.98240}\\[-1pt]{\footnotesize\color{black!65} $n=30$}} & \makecell[r]{0.98093\\[-1pt]{\footnotesize\color{black!65} $n=30$}} & \makecell[r]{0.96904\\[-1pt]{\footnotesize\color{black!65} $n=30$}} & \makecell[r]{0.97488\\[-1pt]{\footnotesize\color{black!65} $n=30$}} & 1\\
\addlinespace[4pt]
$W_1$(speed), m/s $\downarrow$ & \makecell[r]{1.0880\\[-1pt]{\footnotesize\color{black!65} $n=198$}} & \makecell[r]{\textbf{1.0167}\\[-1pt]{\footnotesize\color{black!65} $n=198$}} & \makecell[r]{1.4792\\[-1pt]{\footnotesize\color{black!65} $n=198$}} & \makecell[r]{1.1718\\[-1pt]{\footnotesize\color{black!65} $n=198$}} & 2\\
$W_1(a_{\rm lon})$, m/s$^2$ $\downarrow$ & \makecell[r]{\textbf{0.1407}\\[-1pt]{\footnotesize\color{black!65} $n=198$}} & \makecell[r]{0.1593\\[-1pt]{\footnotesize\color{black!65} $n=198$}} & \makecell[r]{0.1772\\[-1pt]{\footnotesize\color{black!65} $n=198$}} & \makecell[r]{0.1757\\[-1pt]{\footnotesize\color{black!65} $n=198$}} & 1\\
$W_1(a_{\rm lat})$, m/s$^2$ $\downarrow$ & \makecell[r]{\textbf{0.1452}\\[-1pt]{\footnotesize\color{black!65} $n=198$}} & \makecell[r]{0.1645\\[-1pt]{\footnotesize\color{black!65} $n=198$}} & \makecell[r]{0.1687\\[-1pt]{\footnotesize\color{black!65} $n=198$}} & \makecell[r]{0.1706\\[-1pt]{\footnotesize\color{black!65} $n=198$}} & 1\\
$W_1$(jerk), m/s$^3$ $\downarrow$ & \makecell[r]{\textbf{0.0982}\\[-1pt]{\footnotesize\color{black!65} $n=198$}} & \makecell[r]{0.1048\\[-1pt]{\footnotesize\color{black!65} $n=198$}} & \makecell[r]{0.1040\\[-1pt]{\footnotesize\color{black!65} $n=198$}} & \makecell[r]{0.1096\\[-1pt]{\footnotesize\color{black!65} $n=198$}} & 1\\
\bottomrule
\end{tabular*}\endgroup
\end{table*}

\begin{table*}[p]
\centering
\caption{\textbf{nuScenes: 25 metrics for rule-based ego planners.} IDM, STRIVE and PDM-Closed; the fixed 100-scene cohort has 300 rollouts per method. Rates are percentages; CDC-road is a fraction. Motion-distribution scores follow \cref{sec:extended}. Second lines retain counts/denominators or conditional $n$; rollout means use $n=300$ unless shown otherwise. $\mathrm{RM}_{\rm rec}$ uses a group-specific reference pool and the retained-feature definition in \cref{sec:extended}. Heterogeneous-hardware time is unranked; other ranks use unrounded values and shared ranks for ties.}
\label{tab:full_nuScenes_rule}
\begingroup\small\setlength{\tabcolsep}{4pt}\renewcommand{\arraystretch}{1.12}
\begin{tabular*}{\textwidth}{@{\extracolsep{\fill}}lrrrrr@{}}
\toprule
Metric & \textbf{CrashDojo} & SAFE-SIM & CCDiff-DDPM & CCDiff-FM & Our rank\\
\midrule
Ego collision $\uparrow$ & \makecell[r]{\textbf{22.33}\\[-1pt]{\footnotesize\color{black!65} 67/300}} & \makecell[r]{19.33\\[-1pt]{\footnotesize\color{black!65} 58/300}} & \makecell[r]{13.67\\[-1pt]{\footnotesize\color{black!65} 41/300}} & \makecell[r]{10.67\\[-1pt]{\footnotesize\color{black!65} 32/300}} & 1\\
Initial-target contact $\uparrow$ & \makecell[r]{7.33\\[-1pt]{\footnotesize\color{black!65} 22/300}} & \makecell[r]{8.00\\[-1pt]{\footnotesize\color{black!65} 24/300}} & \makecell[r]{\textbf{8.67}\\[-1pt]{\footnotesize\color{black!65} 26/300}} & \makecell[r]{4.67\\[-1pt]{\footnotesize\color{black!65} 14/300}} & 3\\
Target--background $\downarrow$ & \makecell[r]{\textbf{14.00}\\[-1pt]{\footnotesize\color{black!65} 42/300}} & \makecell[r]{14.67\\[-1pt]{\footnotesize\color{black!65} 44/300}} & \makecell[r]{33.67\\[-1pt]{\footnotesize\color{black!65} 101/300}} & \makecell[r]{31.00\\[-1pt]{\footnotesize\color{black!65} 93/300}} & 1\\
ADE (m) $\downarrow$ & \textbf{2.702} & 2.759 & 3.562 & 3.147 & 1\\
FDE (m) $\downarrow$ & \textbf{6.601} & 6.768 & 9.384 & 7.958 & 1\\
Native ORR $\downarrow$ & 3.383 & \textbf{2.884} & 6.701 & 4.778 & 2\\
TTC at $-0.5$ s (s) $\downarrow$ & \makecell[r]{\textbf{0.386}\\[-1pt]{\footnotesize\color{black!65} $n=22$}} & \makecell[r]{0.425\\[-1pt]{\footnotesize\color{black!65} $n=21$}} & \makecell[r]{0.470\\[-1pt]{\footnotesize\color{black!65} $n=21$}} & \makecell[r]{0.414\\[-1pt]{\footnotesize\color{black!65} $n=12$}} & 1\\
\addlinespace[4pt]
Other-pair CR $\downarrow$ & \makecell[r]{\textbf{33.00}\\[-1pt]{\footnotesize\color{black!65} 99/300}} & \makecell[r]{\textbf{33.00}\\[-1pt]{\footnotesize\color{black!65} 99/300}} & \makecell[r]{36.33\\[-1pt]{\footnotesize\color{black!65} 109/300}} & \makecell[r]{35.00\\[-1pt]{\footnotesize\color{black!65} 105/300}} & 1\\
Controlled--background CR $\downarrow$ & \makecell[r]{\textbf{14.00}\\[-1pt]{\footnotesize\color{black!65} 42/300}} & \makecell[r]{14.67\\[-1pt]{\footnotesize\color{black!65} 44/300}} & \makecell[r]{22.33\\[-1pt]{\footnotesize\color{black!65} 67/300}} & \makecell[r]{17.33\\[-1pt]{\footnotesize\color{black!65} 52/300}} & 1\\
Initial-target center off-road $\downarrow$ & \makecell[r]{\textbf{18.100}\\[-1pt]{\footnotesize\color{black!65} 5,430/30,000}} & \makecell[r]{21.753\\[-1pt]{\footnotesize\color{black!65} 6,526/30,000}} & \makecell[r]{29.434\\[-1pt]{\footnotesize\color{black!65} 33,820/114,900}} & \makecell[r]{26.058\\[-1pt]{\footnotesize\color{black!65} 29,941/114,900}} & 1\\
Body off-road, rollout mean $\downarrow$ & \textbf{24.12} & 31.12 & 39.58 & 35.25 & 1\\
\addlinespace[4pt]
First-contact relative speed (m/s) $\uparrow$ & \makecell[r]{1.831\\[-1pt]{\footnotesize\color{black!65} $n=22$}} & \makecell[r]{3.351\\[-1pt]{\footnotesize\color{black!65} $n=24$}} & \makecell[r]{\textbf{4.576}\\[-1pt]{\footnotesize\color{black!65} $n=26$}} & \makecell[r]{2.455\\[-1pt]{\footnotesize\color{black!65} $n=14$}} & 4\\
Reconstructed $\mathrm{RM}_{\rm rec}\downarrow$ & \textbf{0.2456} & 0.2761 & 0.3056 & 0.3148 & 1\\
Mean request time (s) $\downarrow$ & 362.4 & 213.5 & 293.8 & 172.6 & --\\
\addlinespace[4pt]
Mean body area outside $\downarrow$ & \textbf{18.655} & 22.651 & 31.385 & 27.825 & 1\\
Any frame with $>1\%$ outside $\downarrow$ & \makecell[r]{\textbf{31.67}\\[-1pt]{\footnotesize\color{black!65} 95/300}} & \makecell[r]{40.00\\[-1pt]{\footnotesize\color{black!65} 120/300}} & \makecell[r]{88.33\\[-1pt]{\footnotesize\color{black!65} 265/300}} & \makecell[r]{77.67\\[-1pt]{\footnotesize\color{black!65} 233/300}} & 1\\
Body off-road, pooled frames $\downarrow$ & \makecell[r]{\textbf{24.12}\\[-1pt]{\footnotesize\color{black!65} 7,236/30,000}} & \makecell[r]{31.12\\[-1pt]{\footnotesize\color{black!65} 9,336/30,000}} & \makecell[r]{37.61\\[-1pt]{\footnotesize\color{black!65} 43,219/114,900}} & \makecell[r]{33.45\\[-1pt]{\footnotesize\color{black!65} 38,431/114,900}} & 1\\
\addlinespace[4pt]
Wasserstein $\downarrow$ & \makecell[r]{\textbf{0.14443}\\[-1pt]{\footnotesize\color{black!65} $n=891$}} & \makecell[r]{0.15778\\[-1pt]{\footnotesize\color{black!65} $n=891$}} & \makecell[r]{0.16389\\[-1pt]{\footnotesize\color{black!65} $n=891$}} & \makecell[r]{0.16090\\[-1pt]{\footnotesize\color{black!65} $n=891$}} & 1\\
JSD (nats) $\downarrow$ & \makecell[r]{\textbf{0.09918}\\[-1pt]{\footnotesize\color{black!65} $n=891$}} & \makecell[r]{0.10700\\[-1pt]{\footnotesize\color{black!65} $n=891$}} & \makecell[r]{0.10906\\[-1pt]{\footnotesize\color{black!65} $n=891$}} & \makecell[r]{0.10818\\[-1pt]{\footnotesize\color{black!65} $n=891$}} & 1\\
MMD $\downarrow$ & \textbf{0.14027} & 0.14328 & 0.16841 & 0.15427 & 1\\
CDC-road $\uparrow$ & \makecell[r]{\textbf{0.97537}\\[-1pt]{\footnotesize\color{black!65} $n=45$}} & \makecell[r]{0.96549\\[-1pt]{\footnotesize\color{black!65} $n=45$}} & \makecell[r]{0.97367\\[-1pt]{\footnotesize\color{black!65} $n=45$}} & \makecell[r]{0.96797\\[-1pt]{\footnotesize\color{black!65} $n=45$}} & 1\\
\addlinespace[4pt]
$W_1$(speed), m/s $\downarrow$ & \makecell[r]{\textbf{1.4281}\\[-1pt]{\footnotesize\color{black!65} $n=297$}} & \makecell[r]{1.4822\\[-1pt]{\footnotesize\color{black!65} $n=297$}} & \makecell[r]{1.8399\\[-1pt]{\footnotesize\color{black!65} $n=297$}} & \makecell[r]{1.5135\\[-1pt]{\footnotesize\color{black!65} $n=297$}} & 1\\
$W_1(a_{\rm lon})$, m/s$^2$ $\downarrow$ & \makecell[r]{\textbf{0.1771}\\[-1pt]{\footnotesize\color{black!65} $n=297$}} & \makecell[r]{0.1941\\[-1pt]{\footnotesize\color{black!65} $n=297$}} & \makecell[r]{0.2094\\[-1pt]{\footnotesize\color{black!65} $n=297$}} & \makecell[r]{0.2047\\[-1pt]{\footnotesize\color{black!65} $n=297$}} & 1\\
$W_1(a_{\rm lat})$, m/s$^2$ $\downarrow$ & \makecell[r]{\textbf{0.1520}\\[-1pt]{\footnotesize\color{black!65} $n=297$}} & \makecell[r]{0.1670\\[-1pt]{\footnotesize\color{black!65} $n=297$}} & \makecell[r]{0.1757\\[-1pt]{\footnotesize\color{black!65} $n=297$}} & \makecell[r]{0.1690\\[-1pt]{\footnotesize\color{black!65} $n=297$}} & 1\\
$W_1$(jerk), m/s$^3$ $\downarrow$ & \makecell[r]{\textbf{0.1042}\\[-1pt]{\footnotesize\color{black!65} $n=297$}} & \makecell[r]{0.1122\\[-1pt]{\footnotesize\color{black!65} $n=297$}} & \makecell[r]{0.1066\\[-1pt]{\footnotesize\color{black!65} $n=297$}} & \makecell[r]{0.1089\\[-1pt]{\footnotesize\color{black!65} $n=297$}} & 1\\
\bottomrule
\end{tabular*}\endgroup
\end{table*}

\begin{table*}[t]
\centering
\caption{\textbf{Per-planner contact counts of the main comparison.} Rollouts (of 100 per cell) with ego contact with any vehicle (Ego CR) and with the initially selected target (Target), split by ego planner; the planner-class rows of \cref{tab:main_nuPlan,tab:main_nuScenes} are the sums of these cells. Bold marks the highest count per cell.}
\label{tab:per_planner_contact}
\begingroup\footnotesize\setlength{\tabcolsep}{4pt}\renewcommand{\arraystretch}{1.12}
\begin{tabular}{@{}llrrrrrrrr@{}}
\toprule
 & & \multicolumn{4}{c}{Ego CR (count)} & \multicolumn{4}{c}{Target (count)}\\
\cmidrule(lr){3-6}\cmidrule(l){7-10}
Dataset & Planner & CrashDojo & SAFE-SIM & CCDiff-DDPM & CCDiff-FM & CrashDojo & SAFE-SIM & CCDiff-DDPM & CCDiff-FM\\
\midrule
nuPlan & Diffusion Planner & 55 & \textbf{60} & 26 & 30 & 21 & \textbf{26} & 8 & 7\\
nuPlan & Flow Planner & \textbf{60} & 51 & 17 & 20 & \textbf{48} & 43 & 7 & 8\\
nuPlan & IDM & \textbf{66} & \textbf{66} & 46 & 43 & 24 & \textbf{26} & 8 & 6\\
nuPlan & STRIVE & \textbf{59} & 38 & 18 & 19 & \textbf{4} & 1 & \textbf{4} & \textbf{4}\\
nuPlan & PDM-Closed & 52 & \textbf{59} & 42 & 39 & 1 & 2 & \textbf{8} & 2\\
\midrule
nuScenes & Diffusion Planner & \textbf{27} & 23 & 14 & 12 & \textbf{13} & 8 & 8 & 4\\
nuScenes & Flow Planner & \textbf{38} & 23 & 17 & 17 & \textbf{28} & 15 & 11 & 9\\
nuScenes & IDM & \textbf{32} & 30 & 23 & 19 & \textbf{17} & 14 & 14 & 8\\
nuScenes & STRIVE & \textbf{21} & 18 & 9 & 7 & 3 & 5 & \textbf{6} & 4\\
nuScenes & PDM-Closed & \textbf{14} & 10 & 9 & 6 & 2 & 5 & \textbf{6} & 2\\
\bottomrule
\end{tabular}\endgroup
\end{table*}

\begin{table*}[t]
\centering
\caption{\textbf{VLM plausibility screen: the five examples of \cref{fig:qualitative} and three constructed controls.} E is the ego and T1 the selected target. Notes are abridged English translations of the original Chinese reasons; \emph{reasonable} means no identified visible spatiotemporal inconsistency under the supplied evidence. The full 48-case record is provided with the supplementary data.}
\label{tab:vlm_cases}
\begingroup\footnotesize\setlength{\tabcolsep}{3.5pt}\renewcommand{\arraystretch}{1.08}
\begin{tabular}{@{}>{\raggedright\arraybackslash}p{.17\textwidth}>{\raggedright\arraybackslash}p{.16\textwidth}>{\raggedright\arraybackslash}p{.09\textwidth}>{\raggedright\arraybackslash}p{.12\textwidth}>{\raggedright\arraybackslash}p{.40\textwidth}@{}}
\toprule
Request or check & Dataset / planner & Overlap (s) & Label & Reason (abridged)\\
\midrule
Lead braking & nuScenes / IDM & 3.0--9.9 & Reasonable & E and T1 remain visible, with continuous recorded positions during overlap and no identity change.\\
Adjacent-lane cut-in & nuPlan / Flow Planner & 5.2--6.9 & Reasonable$\dagger$ & Both vehicles continue moving during overlap, without a position jump or identity change.\\
Partial intrusion & nuScenes / Flow Planner & 2.7--4.8 & Reasonable & The vehicles continue in similar directions after overlap, with continuous relative-position changes.\\
Crossing & nuPlan / IDM & 3.2--5.0 & Reasonable & E continues moving slowly while T1 moves across the view, with continuous relative motion.\\
Common-lane merge & nuPlan / IDM & 6.0--6.9 & Reasonable & Both vehicles continue moving during overlap, without an identity change or visibility interruption.\\
\midrule
Control: unmodified clip & nuScenes / IDM & None & Reasonable$\dagger$ & E and T1 visible over 0.0--1.5\,s, no box overlap, center distance 13.9$\to$8.0\,m.\\
Control: injected jump & nuPlan / IDM & None & Clearly anomalous & Between 0.7 and 0.8\,s the T1 position jumps 11.9\,m with no visible acceleration or turn.\\
Control: target never rendered & nuPlan / IDM & None & Insufficient evidence & T1 does not appear in any frame, so the interaction with E cannot be observed.\\
\bottomrule
\end{tabular}
\par\vspace{4pt}
\begin{minipage}{\textwidth}\footnotesize
$\dagger$ The full reason also invokes a normal motion range without a supplied threshold (the unsupported-interpretation issue discussed in the text).
\end{minipage}
\endgroup
\end{table*}

\paragraph{Road and event aggregation.}
All event proportions use completed-rollout denominators. Initial-target contact tracks the initially selected set throughout the rollout. A body frame is off-road above a 1\% footprint-area threshold. For rollout $r$, with $N_r$ valid initial-target frames and $B_r$ off-road frames, the rollout mean is $|\mathcal R|^{-1}\sum_r B_r/N_r$ whereas the pooled rate is $\sum_r B_r/\sum_r N_r$. The center metric similarly pools target vehicle-frames, but tests the center point only. The body-area metric averages outside-area fractions within each rollout and then across rollouts. The any-outside metric asks whether any valid frame exceeds 1\%. These metrics include initial discrepancies with the road mask.

\paragraph{Distribution measurements.}
Motion features use the frozen 0.5\,s sampling/differencing protocol. Same-scene histograms compare longitudinal acceleration and lateral acceleration on $[-2,2]$ and jerk on $[-1,1]$, with widths 0.2, 0.2 and 0.1, respectively; speed uses $[0,30]$. The archived binning and support handling are preserved. Wasserstein and JSD average the three acceleration/jerk feature comparisons per eligible rollout; JSD uses natural logarithms. Their scalar averages mix the declared feature scales and should be interpreted alongside the separate physical-unit components. Paired MMD uses the four motion features with actor identities and times jointly valid across methods and recorded reference.

For each of these tables, $\mathrm{RM}_{\rm rec}$ uses the same-domain real trajectories within that planner group to form the common reference histogram, then applies the reconstruction formula below per rollout. CDC-road (coverage of the reference distribution conditioned on road class) groups planner, base road class and feature, requires at least five eligible rollouts per condition, and measures reference-bin mass covered by generated bins at a 2\% support threshold. For learning-based/rule-based groups, it averages 18/27 supported conditions on nuPlan and 30/45 on nuScenes. On nuScenes, Wasserstein and JSD each use 594/891 valid feature comparisons (198/297 rollouts with three finite features); individual feature discrepancies use 198/297 valid rollouts. Other metrics retain their own valid denominators. Further rows of \cref{tab:full_nuPlan_learning,tab:full_nuPlan_rule,tab:full_nuScenes_learning,tab:full_nuScenes_rule} are defined as follows.
\begin{itemize}[leftmargin=1.2em,itemsep=1pt,topsep=2pt]
\item \textit{Other-pair CR} counts rollouts with contact between any vehicle pair other than ego--initial target, including background pairs.
\item \textit{Controlled--background CR} counts contact between method-controlled and uncontrolled non-ego vehicles, classified per frame.
\item \textit{TTC at $-0.5$\,s} is the median time-to-collision queried 0.5\,s before the first new initial-target contact (3\,s horizon, constant relative velocity, oriented separating axis), with $n$ the rollouts whose query returns a finite value.
\item \textit{Native ORR} is each method's own off-road rate over its controlled population; \textit{Mean body area outside} is the body-area metric above.
\item \textit{Mean request time} is measured on heterogeneous hardware and unranked.
\end{itemize}
Contact does not by itself terminate a rollout, so frames after contact remain in the motion and road metrics. Native all-actor ADE/FDE and distribution scores cover all evaluated traffic participants, so they are not selected-target-only fidelity measures.

\paragraph{Target-motion discrepancy for ablations.}
\label{sec:target_motion_metric}
$\mathrm{RM}_{\rm tgt}$ in \cref{tab:ablation_summary} compares the selected target's executed motion with a frozen reference from the development cohort. The target states include initialization and the 7\,s endpoint. Taking every fifth state of the 10\,Hz trajectory gives a grid with $\Delta=0.5$\,s. For position $x_k$ and heading $\psi_k$, define
\begin{equation}
\begin{aligned}
 a_k &= (x_k-2x_{k-1}+x_{k-2})/\Delta^2,\\
 a_k^{\parallel} &= a_k^\top(\cos\psi_k,\sin\psi_k),\\
 a_k^{\perp} &= a_k^\top(-\sin\psi_k,\cos\psi_k),\\
 j_k^{\parallel} &= (a_k^{\parallel}-a_{k-1}^{\parallel})/\Delta.
\end{aligned}
\label{eq:target_motion_features}
\end{equation}
The three features are signed longitudinal acceleration, signed lateral acceleration, and longitudinal jerk in the target frame. Backward differences have no initial padding: acceleration requires three consecutive available grid states, and jerk requires two consecutive valid acceleration samples. Non-finite feature values are excluded.

Acceleration histograms span $[-100,100]$\,m/s$^2$ with width 0.5\,m/s$^2$; jerk spans $[-1000,1000]$\,m/s$^3$ with width 1\,m/s$^3$. Values below the lower edge or at/above the upper edge are clipped into the first or last bin, respectively. Overflow counts are recorded separately, so every finite sample contributes mass. The reference pools recorded nuScenes tracks for every eligible actor in each frozen development scene, counting each scene--actor pair once over the same time window. It is shared across planners, variants, and seeds and contains no generated outcomes.

For each feature $f\in\{a^{\parallel},a^{\perp},j^{\parallel}\}$, normalize the target rollout's counts and the pooled reference counts to bin masses $p_{f,b}$ and $q_{f,b}$. With $B_f$ bins of width $\delta_f$, compute
\begin{equation}
\begin{aligned}
 D_f &= \delta_f\sum_{b=1}^{B_f-1}
 \left|\sum_{i=1}^{b}(p_{f,i}-q_{f,i})\right|,\\
 \mathrm{RM}_{\rm tgt} &= \tfrac{1}{3}\sum_f D_f.
\end{aligned}
\label{eq:target_motion_rm}
\end{equation}
This is the one-dimensional Wasserstein distance between the normalized, equally spaced histograms, averaged over the three features. No feature standardization is applied; the scalar uses the declared physical-unit scales. A rollout contributes only when all three histograms and their references have nonzero mass. Valid rollout scores are averaged within each planner--variant--seed condition; \cref{tab:ablation_summary} reports the mean and sample standard deviation of the three seed summaries. Unlike $\mathrm{RM}_{\rm rec}$ below, $\mathrm{RM}_{\rm tgt}$ uses signed target-frame features, a candidate-actor reference pool, and clipped tail mass.

\paragraph{Reconstructed feature discrepancy.}
We denote the supplementary reconstruction statistic by $\mathrm{RM}_{\rm rec}$ to distinguish it from the target-motion discrepancy $\mathrm{RM}_{\rm tgt}$ in \cref{sec:ablation}. It uses histogram Wasserstein aggregation following CCFM~\cite{li2026ccfm}, with the specific reconstructed features below and a group-specific reference pool. For normalized bin masses $p_f,q_f$ and bin width $\delta_f$,
\begin{equation}
 \mathrm{RM}_{\rm rec}=\frac{1}{3}\sum_{f}\delta_f
 \sum_{b=1}^{B_f-1}\left|\sum_{i=1}^{b}(p_{f,i}-q_{f,i})\right|.
 \label{eq:appendix_rm}
\end{equation}
The reconstruction uses 0.1\,s differences of exported positions, with linear extrapolation at the first sample. With world-frame acceleration $a_k$, its magnitude $A_k=\|a_k\|$, and world heading $\psi_k$, the features are $A_k\cos\psi_k$, $A_k\sin\psi_k$, and $(A_k-A_{k-1})/0.1$. Taking the magnitude before heading projection discards acceleration direction: equal-magnitude acceleration and braking at the same heading can produce identical features. These quantities are therefore not signed body-frame acceleration components.

The first two histograms span $[0,10]$ with width 0.5; the third spans $[0,20]$ with width 1. Values outside these ranges are discarded, and each histogram is normalized over its retained mass. Consequently, $\mathrm{RM}_{\rm rec}$ measures agreement within the retained bins without measuring omitted mass. Reference counts pool corresponding recorded trajectories within the same dataset and planner group; generated counts cover the evaluated actor population. A rollout contributes only when all three comparisons are finite. We interpret this statistic as agreement of the specified reconstruction features, rather than a complete measure of signed motion or physical validity. Its features and reference pool differ from both $\mathrm{RM}_{\rm tgt}$ and the main-table $W_1$/JSD measurements.

\paragraph{Contact-speed condition and missing values.}
Contact speed uses $\|v^a-v^e\|$ from backward 0.1\,s position differences. The reported median is conditioned on a new contact onset with an initially selected target; an initial overlap alone is not a new onset. A non-contact rollout has no contact-speed value, rather than zero, and the conditional support is shown alongside the median.

\paragraph{Reproducibility and operating point.}
The generator configuration uses a Scene Transformer with raster map encoding, 30 history frames, 52 future frames and unicycle dynamics. Evaluation uses 20 flow steps, five candidate samples, target endpoint L-BFGS in the last ten steps, and seed 1. The generator checkpoint SHA-256 is \texttt{aef9b3089312ea978}\allowbreak\texttt{c8162e0656a6cd35}\allowbreak\texttt{23fbfb5cb6c688e6}\allowbreak\texttt{de38d5d12085f5d}. Target identity mappings, provenance of any rerun rollouts, source HDF5 hashes and request configurations are retained in the artifact manifest. In the selected nuPlan scenes, request families comprise 71 longitudinal, 9 cut-in, 14 intrusion/return, 3 alternative-return and 3 crossing requests; nuScenes has 93 longitudinal, 5 cut-in, 1 intrusion/return and 1 crossing request. Together, 164 of the 200 requests (82\%) are longitudinal. No merge request (C5; \cref{sec:knowledge_details}) is in either main cohort; the merge cases of \cref{sec:vlm} are archived rollouts outside the benchmark. These supports do not establish general coverage of every accident class.

\paragraph{Cohort provenance and scope.}
In CrashDojo's nuPlan cohort, all 100 requests have native-map geometry provenance. In nuScenes, 58 have native-map geometry and 42 have fallback geometry from a request builder without native-map boundaries (23 synthetic-current-pose requests, anchored at the actor's current pose, and 19 actor-corridor requests, whose geometry comes from the actor's lane corridor).

The nuPlan results evaluate frozen-generator transfer with target-domain prior and guidance calibration. A 48-case VLM plausibility screen (12 illustrative and 36 stratified cases) is reported in \cref{sec:vlm}.

\paragraph{Per-planner counts and aggregation.}
\Cref{tab:per_planner_contact} splits the Ego CR and initial-target contact counts of \cref{tab:main_nuPlan,tab:main_nuScenes} by ego planner. Each cell contains 100 rollouts, so a count difference equals a percentage-point difference. Against SAFE-SIM, CrashDojo's Ego CR differences on nuPlan are $-5$, $+9$, $0$, $+21$, and $-7$ points in the listed planner order; on nuScenes they are $+4$, $+15$, $+2$, $+3$, and $+4$. The corresponding initial-target differences are $-5$, $+5$, $-2$, $+3$, $-1$ and $+5$, $+13$, $+3$, $-2$, $-3$. These descriptive comparisons explain the heterogeneity behind the class averages. They are not ten independent replications: planners share the same scenes, and the main comparison uses one seed. The CCDiff comparisons also inherit the control-scope difference of \cref{sec:metrics}.

For contact counts $a,b$ on $N$ shared scenes, the number $c$ with contact under both methods can range from $\max(0,a+b-N)$ to $\min(a,b)$. The net difference $a-b$ is fixed, but the number of discordant scene pairs is $a+b-2c$. For nuPlan/Diffusion Planner's Ego CR counts of 55 versus 60, this permits 5--85 discordant pairs; 59 versus 38 for STRIVE permits 21--97. These are feasible count ranges, not observed pairings or uncertainty intervals. The missing overlap prevents recovering a scene-paired confidence interval from the marginal counts alone.

\section{VLM plausibility screen}
\label{sec:vlm}
\begin{table}[t]
\centering
\caption{\textbf{Bounds implied by the three-seed summaries.} Lower bound on Full's difference from each variant over the reported seeds, including display rounding (orientation as in the text). Bold positive entries hold on every seed. -- denotes an unreported input.}
\label{tab:seed_bounds}
\begingroup\footnotesize\setlength{\tabcolsep}{3pt}\renewcommand{\arraystretch}{1.12}
\begin{tabular*}{\columnwidth}{@{\extracolsep{\fill}}lrrrrr@{}}
\toprule
Variant & CR (pp) & MS (m/s) & $\mathrm{RM}_{\rm tgt}$ & Adv.ADE (m) & Adv.FDE (m)\\
\midrule
\multicolumn{6}{@{}l}{\textit{IDM}}\\
A1 & -2.90 & \textbf{4.36} & \textbf{0.0260} & \textbf{0.18} & \textbf{0.55}\\
A2 & -2.86 & \textbf{4.23} & -0.0049 & \textbf{0.11} & -0.03\\
A3 & -7.03 & \textbf{2.68} & \textbf{0.0921} & \textbf{0.67} & \textbf{1.87}\\
A4 & \textbf{4.41} & \textbf{3.72} & \textbf{0.0444} & -- & --\\
\midrule
\multicolumn{6}{@{}l}{\textit{Diffusion Planner}}\\
A1 & \textbf{11.50} & -0.42 & \textbf{0.0245} & \textbf{0.16} & \textbf{0.52}\\
A2 & \textbf{9.47} & -0.24 & -0.0211 & \textbf{0.09} & -0.14\\
A3 & \textbf{1.90} & -0.56 & \textbf{0.0729} & \textbf{0.17} & \textbf{1.62}\\
A4 & \textbf{11.47} & \textbf{0.33} & \textbf{0.0113} & -- & --\\
\bottomrule
\end{tabular*}\endgroup
\end{table}

A generated contact is informative about the planner only if the executed interaction is spatiotemporally continuous. We therefore screen executed rollouts for rendering or tracking artifacts, such as position jumps, identity changes, or a target that never appears, with a fixed VLM whose every observation is tied to a vehicle, a time interval, and a supplied measurement. The auxiliary VLM descriptions referenced in the main text are the outputs of this screen.

\paragraph{Protocol.}
Qwen3-VL-32B-Instruct (BF16, greedy decoding, at most 512 new tokens; revision \texttt{0cfaf48\allowbreak 183f594c31}) receives one MP4 per rollout with all recorded frames at 10\,Hz (100 for nuScenes, 70 for nuPlan) and a video-pixel budget of $4.096\times10^7$. The fixed world view shows timestamps, the ego E (blue), the target T1 (orange), surrounding vehicles (gray), and the native-map drivable mask. Code supplies recorded-box overlap intervals, per-vehicle visible-frame counts, and the maximum adjacent-frame displacement of E and T1. Because these measurements are inputs, a matching explanation shows consistency with supplied information, not independent recovery from video. A fixed Chinese instruction (SHA-256 prefix \texttt{42bcf582}) requests a label and a one-line reason naming the time and vehicle. \emph{Clearly anomalous} requires a locatable spatiotemporal contradiction; a target with no visible frame requires \emph{insufficient evidence}; \emph{reasonable} means no identified visible inconsistency. Collision alone is not an anomaly. The instruction forbids judging displacement against an unsupplied threshold and requests no mechanism-match or responsibility judgment; method names are withheld.

\paragraph{Coverage and outcome.}
Twelve illustrative rollouts (the five examples of \cref{fig:qualitative} plus seven further scenes, five planners) are complemented by a 36-case extension selected before inference by hash order within each request family and round-robin over road classes (family counts 8/8/8/2/4/6; 28 nuPlan, 8 nuScenes). The 48 rollouts contain 15 target contacts. All 48 receive \emph{reasonable}, i.e.\ no discontinuity, identity change, or missing target is reported. Three constructed controls show that every label is reachable: an injected 11.9\,m jump of T1 is flagged as \emph{clearly anomalous} between 0.7 and 0.8\,s, a never-rendered target is labelled \emph{insufficient evidence}, and an unmodified clip is labelled \emph{reasonable}. \Cref{tab:vlm_cases} lists the five examples of \cref{fig:qualitative} and the controls; the full 48-case record, including the extension roster, is provided with the supplementary data (\texttt{data/vlm/}).

\paragraph{Explanation checks and limits.}
Reading the reasons against the supplied numbers exposes two kinds of explanation issue: one numerical slip (a nuPlan/Diffusion Planner partial-intrusion reason states 0.67\,m where the supplied maximum is 0.698\,m) and unsupported interpretation in 18 of the 36 extension reasons, which invoke a ``normal'' range although no threshold was supplied. These are explanation errors, not physically invalid trajectories, and they show that positive labels cannot replace the numerical cross-check. The screen has no independent human labels, and the supplied statistics also reveal the constructed anomalies, so it does not isolate the video's contribution. It complements, and does not replace, the trajectory-based checks of \cref{sec:extended}.

\section{Consistency of the reported seed summaries}
\label{sec:seed_consistency}
\Cref{tab:ablation_summary} reports the mean $\bar x$ and sample standard deviation $s_x$ of three seed summaries per cell. Because the deviations $x_i-\bar x$ sum to zero, Cauchy--Schwarz gives $|x_i-\bar x|\le\frac{n-1}{\sqrt n}s_x$, so any two cells satisfy
\begin{equation}
 \min_{i,j}(x_i-y_j)\geq(\bar x-\bar y)-\frac{n-1}{\sqrt n}(s_x+s_y),
 \label{eq:seed_difference_bound}
\end{equation}
with factor $2/\sqrt3$ for three seeds and no covariance or normality assumption. We widen every printed value by half a unit in its last decimal place, orient CR and MS as Full minus variant and the discrepancy/displacement metrics as variant minus Full, and round the bounds in \cref{tab:seed_bounds} downward. A positive bound means the comparison has the same direction on every reported seed; a nonpositive bound leaves it unresolved. The source package includes a script that parses \cref{tab:ablation_summary} and regenerates this table.

For both planners, Full has lower $\mathrm{RM}_{\rm tgt}$, Adv.ADE, and Adv.FDE than A3 on every reported seed ($\mathrm{RM}_{\rm tgt}$ bounds 0.0921 and 0.0729). Full's CR advantage over A4 holds for both planners; over A1, A2, and A3 it holds for Diffusion Planner (1.90 percentage points against A3), while the IDM summaries leave these directions unresolved, as they do for the small Full--A2 $\mathrm{RM}_{\rm tgt}$ differences.

\end{document}